\documentclass[acmlarge]{acmart}

\setcopyright{acmcopyright}
\copyrightyear{2018}
\acmYear{2018}
\acmDOI{10.1145/1122445.1122456}

\acmJournal{POMACS}
\acmVolume{37}
\acmNumber{4}
\acmArticle{111}
\acmMonth{8}

\UseRawInputEncoding

\usepackage{centernot}

\usepackage{amsmath,amsthm, amssymb}
\usepackage{graphicx}
\usepackage{epstopdf}
\usepackage{float}
\usepackage{enumerate}
\usepackage{multicol}
\usepackage{algorithmic}
\usepackage[lined,boxed,commentsnumbered,linesnumbered,ruled]{algorithm2e}

\usepackage{pbox}
\usepackage{multirow}
\usepackage{hhline}
\usepackage{color}
\usepackage{gensymb}
\usepackage{subfigure}

\newcommand{\nop}[1]{}

\newtheorem{definition}{Definition}
\newtheorem{theorem}{Theorem}

\newtheorem{lemma}{Lemma}

\SetKwInput{KwInput}{Input}
\SetKwInput{KwOutput}{Output}
\let\oldKwInput\KwInput
\renewcommand{\KwInput}[1]{%
  \makebox[\widthof{\KwOutput{}}][l]{\oldKwInput{}#1}%
}

\newcommand{\indep}{\mathrel{\text{\scalebox{1.07}{$\perp\mkern-10mu\perp$}}}}

\newcommand{\nindep}{\centernot{\indep}}

\begin{document}
\begin{sloppy}

\title{Feature Selection for Efficient Local-to-Global Bayesian Network Structure Learning}
\author{Kui Yu, Zhaolong Ling}
\email{yukui@hfut.edu.cn}
\email{z\_dragonl@163.com}
\affiliation{%
\city{School of Computer Science and Information Engineering}
  \institution{Hefei University of Technology}
  \streetaddress{485 Daxia Road, Shushan District}
  \city{Hefei}
\postcode{230601}
   \country{China}
  }

\author{Lin Liu}
\email{Lin.Liu@unisa.edu.au}
\affiliation{%
\city{UniSA STEM}
 \institution{ University of South Australia}
 \streetaddress{Mawson Lakes Blvd, Mawson Lakes}
\city{Adelaide}
\postcode{5095}
  \state{SA}
 \country{Australia}
}

\author{Hao Wang}
\email{jsjxwangh@hfut.edu.cn}
\affiliation{%
\city{School of Computer Science and Information Engineering}
  \institution{Hefei University of Technology}
  \streetaddress{485 Daxia Road, Shushan District}
  \city{Hefei}
\postcode{230601}
   \country{China}
  }

\author{Jiuyong Li}
\email{Jiuyong.Li@unisa.edu.au}
\affiliation{%
\city{UniSA STEM}
 \institution{ University of South Australia}
 \streetaddress{Mawson Lakes Blvd, Mawson Lakes}
\city{Adelaide}
\postcode{5095}
  \state{SA}
 \country{Australia}
}


\begin{abstract}
Local-to-global learning approach plays an essential role in Bayesian network (BN) structure learning. Existing local-to-global learning algorithms first construct the skeleton of a DAG (directed acyclic graph)  by learning the MB (Markov blanket) or PC (parents and children) of each variable in a data set, then orient edges in the skeleton. However, existing MB or PC learning methods are often computationally expensive especially with a large-sized BN, resulting in inefficient local-to-global learning algorithms. To tackle the problem, in this paper, we develop an efficient local-to-global learning approach using feature selection. Specifically, we first analyze the rationale of the well-known Minimum-Redundancy and Maximum-Relevance (MRMR) feature selection approach for learning a PC set of a variable. Based on the analysis, we propose an efficient F2SL (feature selection-based structure learning) approach to local-to-global BN structure learning.  The F2SL approach first employs the MRMR approach to learn a DAG skeleton, then orients edges in the skeleton. Employing independence tests or score functions for orienting edges, we instantiate the F2SL approach into two new algorithms, F2SL-c (using independence tests) and F2SL-s (using score functions).
Compared to the state-of-the-art local-to-global BN learning algorithms, the experiments validated  that the proposed algorithms in this paper are more efficient and provide competitive structure learning quality than the compared algorithms.
\end{abstract}

\keywords{Bayesian network, Feature selection, Local-to-global structure learning, Markov blanket}

\maketitle

\section{Introduction}\label{sec1}

Learning Bayesian network (BN)  from observational data is an important problem in data mining, playing an essential part in inferring conditional independence and causal relations between variables~\cite{aliferis2010local1}. BN learning has had many applications in various areas such as bioinformatics~\cite{friedman2000using}, neuroscience~\cite{bielza2014bayesian},  and information retrieval~\cite{de2004bayesian}.

The structure of a BN is represented by a DAG (directed acyclic graph) where nodes of the DAG represent the variables and edges represent dependence between variables. When there is an edge $X\rightarrow Y$, $X$ is known as a parent of $Y$ and $Y$ is a child of $X$.
In general, learning a DAG over a large number of variables is computationally intractable~\cite{chickering2004large,scutari2019learning}.
To alleviate the computational complexity,  the local-to-global approach was proposed to reduce the DAG search space~\cite{margaritis2000bayesian,tsamardinos2006max,gao2017local}.
Instead of searching the entire DAG space over all the variables simultaneously, the local-to-global approach first finds the Markov blanket (MB) or parents and children (PC) of each variable (without distinguishing parents from children), then uses the learnt MB (or PC) set of each variable to construct the DAG skeleton, and finally orients edges in the skeleton.  An MB of a node in a BN consists of the parents, children, and spouses (i.e. other parents of the node's children) of the node in a BN.
Figure~\ref{fig1-1} gives an example of a MB in the BN of lung cancer~\cite{guyon2007causal}. The MB of \emph{Lung cancer} consists of \emph{Smoking} and \emph{Gentics} (parents), \emph{Coughing} and \emph{Fatigue} (children), and \emph{Allergy} (spouse).

\begin{figure}[t]
\centering
\includegraphics[height=1.4in,width=3in]{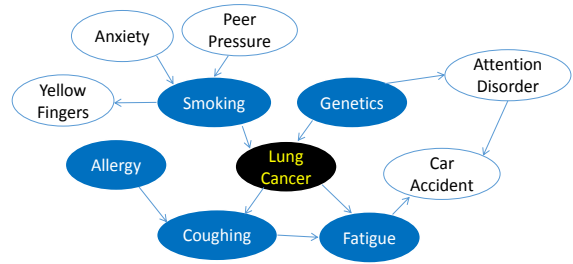}
\caption{An example of a MB in a lung-cancer Bayesian network}
\label{fig1-1}
\end{figure}

The current MB learning methods are mainly categorized into two types: simultaneous MB learning and divide-and conquer MB learning.
However, these methods are either inefficient or ineffective.  Given a variable of interest, simultaneous MB learning algorithms learn the MB of the variable simultaneously without distinguishing PC from spouses and are not capable of constructing high quality DAG skeletons especially when a data set has a small number of data samples with high-dimensionality, but they are computationally efficient. Divide-and conquer methods first employ a Parent-Child (PC) learning algorithm to learn the PC set of the variable, then find the spouses of the variable. They are effective for DAG skeleton construction, but they are not computationally efficient when the size of PC is large. This implies  that existing local-to-global BN structure learning algorithms are either inefficient or ineffective depending on which type of MB or PC learning methods are used.

Feature selection aims to select a subset of features with regard to a class variable of interest from the original set of features,  and it is an essential preprocessing step for model building or data understanding in data analytics~\cite{brown2012conditional}. Existing feature selection methods can be broadly categorized into  filtering, wrapper, and embedded methods~\cite{li2017feature,yu2020causality}. Filter methods are classifier independent,  and the other two types of methods are classifier dependent.

Filtering feature selection methods have been attracting major attention, due to their fast processing speed and independence of prediction models.
Studies have shown that under certain assumptions, the MB of a class variable is the optimal feature set for supervised machine learning tasks, while existing filtering feature selection methods attempt to find an approximate MB (e.g. PC) of a class variable~\cite{tsamardinos2003towards}.
A question is whether we can make use of efficient filtering feature selection methods to learn the MB/PC set of each variable in a dataset to significantly improve computational efficiency of BN structure learning.
To investigate this question, in this paper, we first establish the theoretical link of filtering feature selection methods with local BN structure learning (i.e., MB/PC learning) and then propose an efficient F2SL (\underline{F}eature \underline{S}election-based \underline{S}tructure \underline{L}earning) approach to speeding up local-to-global BN structure learning.  Our contributions can be summarized as follows.


\begin{itemize}
\item  We employ a well-known mutual information-based feature selection approach, called Min-Redundancy and Maximum-Relevance (MRMR)~\cite{brown2012conditional}, to find parents and children for speeding up structure learning. To answer why the MRMR approach is able to learn the PC of a variable of interest, we first analyze the relationship of the objective function of MRMR and MB learning. Then we analyze the output of MRMR related to the MB. And finally we discuss the instantiation of MRMR for PC learning.

\item Based on the analysis above, we propose the F2SL approach to local-to-global BN structure learning.  F2SL employs the MRMR approach to learn the PC of each variable in a data set, then constructs the DAG skeleton, and finally it orients edges in the skeleton. Using independence tests or score functions for orienting edges, F2SL is instantiated to two algorithms, F2SL-c (using independence tests) and F2SL-s (using score functions).

\item We conduct extensive experiments to validate  F2SL-c and  F2SL-s against the five representative local-to-global BN learning algorithms. The experiments results show that the proposed algorithms are significantly faster and also achieve better structure learning quality than the five rivals.
\end{itemize}

The paper is organized as follows. Section~\ref{sec2} reviews the related work, and Section~\ref{sec3} gives notations and definitions. Section~\ref{sec4} presents an analysis in theory, while Section~\ref{sec5} proposes the new algorithms. Section~\ref{sec6} describes and discusses the experiments and Section~\ref{sec7} concludes the paper.

\section{Related work}\label{sec2}

In the past decades, learning BN structures has been an important task in data mining and machine learning~\cite{heckerman2008tutorial,koller2009probabilistic}. There are two main types of structure learning methods: score-based and constraint-based methods~\cite{zhang2018learning,glymour2019review}.
Score-based algorithms use a scoring function to perform a global structure learning over a search space of possible DAGs over all the variables in a data set~\cite{chickering2002optimal,de2018entropy}.  Constraint-based methods employ independence tests to first estimate whether there is an edge between two variables and then orient edge directions~\cite{cheng2002learning,colombo2014order}.
Existing BN learning approaches  formulated the BN structure learning problem as a traditional combinatorial optimization problem and depend on various local heuristics for enforcing the acyclicity constraint.

To avoid the combinatorial constraint, Zheng et al.~\cite{zheng2018dags}  have formulated the BN structure learning problem as a continuous optimization problem instead of the traditional combinatorial optimization problem.  Some recent studies have leveraged the idea in~\cite{zheng2018dags} to learn BN structures using deep neutral networks. Yu et al.~\cite{yu2019dag} have designed a BN structure learning algorithm using graph neural networks and Zhang et al.~\cite{zhang2019d}  have proposed a variational autoencoder-based method for learning BN structures.

Since the search space of DAGs is combinatorial and exponential with the number of variables, existing global BN structure learning methods are often computationally infeasible when the number of variables is large. Then to improve efficiency of BN structure learning, local-to-global BN structure learning methods were proposed which contain two steps: skeleton learning and edge orientation. In the skeleton learning step, the local-to-global approach first learns the MB or PC of each variable in a dataset independently, then constructs the DAG skeleton (i.e. the undirected graph) using the learnt MB or PC sets. Through learning each variable's MB or PC locally, the local-to-global approach significantly  reduces the potential DAG search space, and thus can be scalable to thousands of variables. In the edge orientation step, edges are oriented in the skeleton using independence tests or score functions.

How to efficiently learn the MB or PC of a variable for skeleton learning is the key to existing local-to-global BN learning algorithms.
To learn skeletons, many MB and PC learning algorithm have been proposed and they fall in two types: constraint-based methods and score-based approaches. Constraint-based MB learning algorithms employ independence tests and are mainly divided into two types: simultaneous learning approach and divide-and conquer approach.  The representative algorithms of the former type include GSMB~\cite{margaritis2000bayesian}, IAMB~\cite{tsamardinos2003algorithms}, and Inter-IAMB~\cite{tsamardinos2003algorithms}, while the representative algorithms of the latter type are
 HITON-MB~\cite{aliferis2003hiton,aliferis2010local1}, MMMB~\cite{tsamardinos2003time}, PCMB~\cite{pena2007towards}, STMB~\cite{gao2016efficient}, and BAMB~\cite{ling2019BAMB}. PC-simple~\cite{li2015practical}, MMPC~\cite{tsamardinos2003time} and HITON-PC~\cite{aliferis2003hiton}  are the three widely used algorithms for learning  PC of a variable using independence tests. Score-based MB (or PC) learning methods use a score function to learn MB or PC, and mainly includes the SLL (Score-based Local Learning)~\cite{niinimki2012local}  and S$^2$TMB (Score-based Simultaneous MB) algorithms~\cite{gao2017efficient2}.

Based on these MB or PC learning algorithms, several local-to-global structure learning methods were proposed.
The GSBN~\cite{margaritis2000bayesian} algorithm employs the GSMB algorithm for learning skeleton and orients edge  using independence tests.  SLL+C~\cite{niinimki2012local} uses the SLL algorithm to learn the MB of each variable and employs independence tests for orienting edges while SLL+G~\cite{niinimki2012local} uses score-based methods for edge orientations.
MMHC~\cite{tsamardinos2006max} uses MMPC for learning skeletons and employs a score function and hill-climbing search strategy for orienting edges. Thus both SLL+C and MMHC belong to a hybrid local-to-global approach.
Gao et al.~\cite{gao2017local} recently proposed a novel GGSL (Graph Growing Structure Learning) algorithm for local-to-global structure learning. Instead of finding the MB of each variable in advance, GGSL first randomly selects a variable and learns the local structure around the variable using the S$^2$TMB algorithm, then iteratively applies the local learning procedure to the variable's neighbors for gradually expanding the learned local BN structure until  a global BN structure is achieved.


Existing MB or PC learning algorithms have the following main drawbacks. GSMB and IAMB are efficient, but the number of samples required by them grows exponentially with the size of the MB of the target variable, since they use the entire set of variables selected currently as conditioning sets for independence tests. HITON-MB, MMMB, PCMB, STMB, and BAMB mitigate the problem of the large sample requirement by performing an exhaustive subset search within the variables selected currently, but the search is computationally expensive when the size of the currently selected variables becomes large. The computational cost of SLL and S$^2$TMB is determined by BN structure learning algorithms, since at each iteration SLL and S$^2$TMB need to use a  BN structure learning algorithm to learn a local BN structure (involving all variables selected currently) around a variable. Due to the computational complexity of existing BN structure learning algorithms, they face the scalability issues when the size of the local BN structure  becomes large.
To improve the MB learning efficiency, Pellet et al.~\cite{pellet2008using} proposed the TC (Total Conditioning) algorithm for learning BN structures from Gaussian data.  The TC algorithm employs a regression-based feature-selection method to learn DAG skeletons and then uses independence tests to orient edges.

In the past decade, researchers have proposed many methods for distinguishing causes from effects purely from observational data in the two-variable case~\cite{zhang2012identifiability}. These methods are mainly divided into two types: methods based on additive noise models~\cite{shimizu2006linear,hoyer2009nonlinear} and methods based on information geometric causal inference~\cite{janzing2012information}. The focus of this paper is on learning a complete BN structure containing all variables in a problem of interest. For readers who are interested in the research on distinguishing causes from effects in the two-variable case, more references can be found in the recent survey  proposed by Mooij et al~\cite{mooij2016distinguishing}.

In this paper, we focus on address the computational problem of existing MB or PC learning algorithms and focus on learning BN structures from discrete data with multivariate random variables.

\section{Notations and definitions}\label{sec3}
\subsection{Bayesian network and Markov blanket}\label{sec31}
In this section, we will introduce some basic definitions.
Let $P$ be the joint probability distribution represented by a DAG $G$  over  a set of random variables
 $V=\{V_1,\cdots, V_M\}$.
We use $V_i\indep V_j|S$ to denote that $V_i$  and $V_j$ are conditionally independent given
$S\subseteq V\setminus\{V_i, V_j\}$, and $V_i\nindep V_j|S$ to represent that $V_i$  and $V_j$ are conditionally dependent given $S$.
 The definition of conditional independence (and dependence) is given as follows.
\begin{definition}[Conditional independence] Given two distinct variables $V_i, V_j\in V$  are said to be conditionally independent given a subset of variables $S\subseteq V\setminus\{V_i, V_j\}$ (i.e., $V_i\indep V_j|S$), if and only if $P(V_i,V_j|S)=P(V_i|S)P(V_j|S)$. Otherwise, $V_i$ and $V_j$ are conditionally dependent given $S$, i.e., $V_i\nindep V_j|S$.
\label{def2-1}
\end{definition}
The symbols $pa(V_i)$, $ch(V_i)$, and $sp(V_i)$ denote the sets of parents, children, and spouses of $V_i$, respectively.
We call the triplet $\langle V, G, P\rangle$ a \emph{Bayesian network (BN)} if $\langle V, G, P\rangle$ satisfies the Markov condition: every variable is independent of any subset of its non-descendant variables given its parents in $G$~\cite{pearl2014probabilistic}. In a BN $\langle V, G, P\rangle$, by the Markov condition, the joint probability $P$ can be decomposed into the product of conditional probabilities as
\begin{equation}\label{eq3-1}
\small
P(V_1, V_2,\cdots,V_M) = \prod_{i=1}^M{P(V_i|pa(V_i))}.
\end{equation}


\begin{definition}[d-separation~\cite{pearl2014probabilistic}] In a DAG $G$, a path $\pi$ is said to be d-separated (or blocked) by a set of vertices $S\subset V$ if and only if (1)
$\pi$ contains a chain $V_i\rightarrow V_k\rightarrow V_j$ or a fork $V_i\leftarrow V_k\leftarrow V_j$ such that the middle vertex $V_k$ is in $S$, or
(2) $\pi$ contains an inverted fork (or collider) $V_i\rightarrow V_k\leftarrow V_j$ such that the middle vertex $V_k$ is not in $S$ and such that no descendant of $V_k$ is in $S$.

A set $S$ is said to d-separate $V_i$ from $V_j$ if and only if $S$ blocks every path from a vertex in $V_i$ to a vertex in $V_j$.
\label{def3-0}
\end{definition}

\begin{definition}
[Faithfulness]~\cite{spirtes2000causation} Given a BN $\langle V,G,P\rangle$, $P$ is faithful to $G$ if $\forall V_i,\ V_j\in V$, $\exists S\subseteq V\setminus\{V_i, V_j\}$ d-separates  $V_i$ and $V_j$ in $G$ if  $V_i\indep V_j|S$ holds in $P$.
\label{def3-2}
\end{definition}

The faithfulness assumption establishes a relation between a probability distribution $P$ and its underlying DAG $G$.
In a BN, the faithfulness assumption implies that two variables $V_i,\ V_j\in V$ that are d-separated with each other by a subset $S\subseteq F\setminus\{F_i, F_j\}$ in $G$ are conditionally independent conditioning on $S$ in $P$.
Under the assumption, we can use conditional independence tests, instead of d-separation, to find all dependencies or independencies entailed with a Bayesian network.

\begin{definition}
[Markov blanket]\cite{pearl2014probabilistic}
Under the faithfulness assumption, the MB of a variable $V_i$ in $G$, noted as $MB(V_i)$, is unique and consists of parents, children and spouses of $V_i$ .
\label{def3-3}
\end{definition}

In the following, Lemma~\ref{lem3-1} states the dependencies between a variable and its parents (or children).  Lemma~\ref{lem3-2} denotes the independence/dependence relations of a variable and its spouses.

\begin{lemma}~\cite{spirtes2000causation}
Under the faithfulness assumption, for $V_j\in V$ and $V_i\in V$, there is an edge between $V_j$ and $V_i$ if and only if $V_j\nindep V_i|S$, for all $S\subseteq V\setminus\{V_j, V_i\}$.
\label{lem3-1}
\end{lemma}
Lemma~\ref{lem3-1} indicates that if $V_j$ is a parent or a child of $V_i$, $V_j$ and $V_i$ are conditionally dependent given any subset $S$ of $V\setminus\{X, V_i\}$.

\begin{lemma}\cite{spirtes2000causation}
In a Bayesian network, assuming that $V_i$ is adjacent to $V_k$, $V_j$ is adjacent to $V_k$, and $V_i$ is not adjacent to $V_j$ (e.g. $V_i\rightarrow V_k\leftarrow V_j$), if
$\forall S\subseteq V\setminus\{V_i, V_k, V_j\}$, $V_i\indep V_j|S$ and $V_i\nindep V_j|S\cup \{V_k\}$ hold, then $V_j$ is a spouse of $V_i$.
\label{lem3-2}
\end{lemma}

Lemma~\ref{lem3-2} states that if $V_j$ is a spouse of $V_i$, $V_j$ and $V_i$ are independent conditioning on a subset $S$ excluding their common child, but they are dependent conditioning on $S$ including their common child.

\subsection{Mutual information and conditional independence}\label{sec32}
Given variable $X$, the entropy of $X$ is defined as
\begin{equation}\label{eq_2}
H(X)=-\Sigma_{x}P(x)\log{P(x)}
\end{equation}

The entropy of $X$ after observing values of another variable $Y$ is defined as
\begin{equation}\label{eq_3}
H(X|Y)=-\Sigma_{y}P(y)\Sigma_{x}P(x|y)\log{P(x|y)}.
\end{equation}

In Eq.(\ref{eq_2}) and Eq.(\ref{eq_3}), $P(x)$ is the prior probability of $X=x$ (i.e. the value $x$ that $X$ takes), and $P(x|y)$ is the posterior probability of $X=x$ given  $Y=y$.
In order to calculate  Eq.(\ref{eq_2}) we need the estimated distributions $P(x)$ and $P(x|y)$.
For a discrete dataset with $N$ independently and identically distributed samples, we estimate the probability of $P(X=x)$ using maximum likelihood, i.e. the frequency of $X=x$ occurring in the dataset divided by the total number of data samples (i.e. $N$) of the dataset.  The posterior probability $P(x|y)$ is calculated by the frequency of $X=x$ occurring given that $Y=y$ holds in the dataset divided by the total number of data samples $N$.

According to Eq.(\ref{eq_2}) and Eq.(\ref{eq_3}), the mutual information between $X$ and $Y$, denoted as $I(X,Y)$, is defined as
\begin{equation}\label{eq_4}
\begin{array}{rcl}
I(X; Y)&=&H(X)-H(X|Y)\\
&=&\Sigma_{x, y}P(x,y)\log\frac{P(x,y)}{P(x)P(y)}.
\end{array}
\end{equation}

From Eq.~(\ref{eq_4}), the conditional mutual information between $X$ and $Y$ given another variable $Z$ is defined as:
\begin{equation}\label{eq_5}
\begin{array}{rcl}
 I(X;Y|Z)& =& H(X|Z)-H(X|YZ) \\
 &=&\Sigma_{z\in Z}P(z)\Sigma_{x\in X, y\in Y}P(x,y|z)\log\frac{P(x,y|z)}{P(x|z)P(y|z)}.
\end{array}
\end{equation}

In Eq.(\ref{eq_4}) and Eq.(\ref{eq_5}),  i$I(X;Y)=0$  if $X$ and $Y$ is independent (i.e., $P(X,Y)$=$P(X)P(Y)$), $I(X;Y)=0$ and  $I(X;Y|Z)=0$ if $X$ and $Y$ is conditionally independent given $Z$.

\section {MRMR feature selection \& MB learning}\label{sec4}

Given a data set $D$ consisting of a variable set $V$ and  the class attribute  $C\in V$, in this section, we will analyze the output of the MRMR feature selection approach with relation to the MB of  $C$.

\subsection{Objective functions of MRMR and MB learning}\label{sec41}

\textbf{Objective function of MB learning.}
Under the faithfulness assumption, $MB(C)$ is the optimal feature set for the classification problem with $C$ as the class variable, and $MB(C)$ satisfies the following property~\cite{yu2019MCFS}.

\begin{theorem}
$\forall S\subset V\setminus C$, $I(C;MB(C))\geq I(C;S)$.
\label{the4-1}
\end{theorem}
In Theorem~\ref{the4-1}, except for $S=MB(C)$, if $S$ is a superset of $MB(C)$,  $I(C;MB(C))= I(C;S)$ also holds. In~\cite{yu2019MCFS},  the authors do not consider this case.  This case means that although by the property of the MB,  $C$ is conditionally independent of the remaining variables $S$ conditioning on $MB(C)$ (i.e., $\forall S\subseteq F\setminus MB(C),\ P(C|S,MB(C))=P(C|MB(C)$),  $C$ and $S$ also conditionally independent conditioning on a  superset of $MB(C)$, i.e., $\forall S, S'\subset F\setminus MB(C),\ P(C|S,MB(C)\cup S')=P(C|MB(C)\cup S')$. Then Theorem~\ref{the4-1} indicates that learning $MB(C)$ is equivalent to finding a subset $S\subseteq V\setminus C$ that maximizes $I(S;C)$ and  $MB(C)$ is the minimal and optimal set of $S$, since adding features to the $MB(C)$ set does not increase the mutual information to $C$.  Although the issue of the superset exists in Theorem~\ref{the4-1}, it does not put any impact on  existing algorithms for learning  $MB(C)$ or $PC(C)$.  Since the MB property  $\forall S\subseteq F\setminus MB(C),\ P(C|S,MB(C))=P(C|MB(C))$) holds,  these algorithms  can use this property to remove redundant features in the learnt MB and are able to  identify the minimal MB of  $C$ (i.e.,  parents, children, and spouses of $C$). In addition,  if we assume that all conditional independence tests are reliable and  the faithfulness assumption holds, existing MB learning algorithms achieve a correct MB of a target variable in theory.

However, in Theorem~\ref{the4-1},  it  is a challenging combinatorial optimization problem for identifying $MB(C)$ from $F$.
Therefore, almost all MB learning methods have adopted a greedy strategy by considering features one by one to find $S^*$. Specifically, at each iteration, given the currently selected set of features, $S$, they choose the feature $X$ in $V\setminus (S\cup \{C\})$ that maximizes $I((S\cup \{X\});C)$. Since $I((S\cup X);C)=I(S;C)+I(X;C|S)$ holds and at each iteration $I(S;C)$ is the same for each variable $X$, this greedy strategy can be formulated as Eq.(\ref{eq6}) as follows.

\begin{equation}
X^*=\mathop{\arg\max}_{X\in V\setminus (S\cup \{C\})}I(X;C|S).\\
\label{eq6}
\end{equation}

To solve Eq.(\ref{eq6}), there are two types of MB learning methods. One is to calculate $I(X;C|S)$ in Eq.(\ref{eq6})  using the entire set $S$ of features currently selected, i.e., all the features in $S$,
such as GSMB~\cite{margaritis2000bayesian} and IAMB~\cite{tsamardinos2003algorithms},  while the other is to calculate $I(X;C|S)$ using all possible subsets of $S$, such as HITON-MB~\cite{aliferis2003hiton} and MMMB~\cite{tsamardinos2003time}.
When the size of $S$ increases, it will be impractical for the first type of methods to compute $I(X;C|S)$ because the computation demands a large number of training samples while it will be computationally intractable for the second type of methods since the number of subsets of $S$ will be exponential to the number of features in $S$.

\textbf{Objective function of MRMR.}
To tackle these problems, since $I(X;S;C)=I(X;S)-I(X;S|C)=I(X;C)-I(X;C|S)$ holds, MRMR decomposes Eq.(\ref{eq6}) into  Eq.(\ref{eq7}) as follows.
\begin{equation}\label{eq7}
X^*=\mathop{\arg\max}_{X\in V\setminus S}\{I(X;C)-I(X;S)+I(X;S|C)\}.
\end{equation}
In Eq.(\ref{eq7}), $I(X;C)$ represents the relevancy of $X$ to $C$, $I(X;S)$ denotes the redundancy of $X$ with respect to $S$, and
$I(X;S|C)$ indicates the class-conditional relevance, which considers the situation where a feature $X$ provides more predictive information by jointly with other features (i.e., those in $S$) than by itself with respect to $C$.

When the size of $S$ becomes large, if we direct compute $I(X;S)$ and $I(X;S|C)$ in Eq.(\ref{eq7}),  on the one hand,  it will result in computationally expensive; on the other hand, the big size of $S$ will lead to the large sample requirement to guarantee the reliable results of computing  $I(X;S)$ and $I(X;S|C)$.

To reduce high-order mutual information computations to low-order mutual information calculations, we require highly restrictive assumptions made on the dependence/independence between features.
To deal with the class-conditional relevance term $I(X;S|C)$,  Assumption 1 is the assumption by a naive Bayes classifier and it assumes that the features in $V\setminus C$ are pairwise independent conditioning on $C$.
By Assumption 1, we get $I(X;S|C)=0$ and the term $\{I(X;C)-I(X;S)+I(X;S|C)\}$ in Eq.(\ref{eq7}) is rewritten as $\{I(X;C)-I(X;S)\}$.
This assumption makes MRMR only interested in which variables has an edge with $C$, i.e., parents and children of  $C$.

\textbf{Assumption 1.} $\forall V_i, V_j\in V$ and $i\neq j$, $V_i$ and $V_j$ are assumed to be conditionally independent given the class attribute $C$, that is, $P(V_i,V_j|C) = P(V_i|C)P(V_j|C)$.

To tackle the redundancy term  $I(X;S)$, Assumption 2 assumes
the selected features in $S$ are conditionally independent with each other conditioning on a feature $X$ in $V\setminus S$ (i.e. conditioning on a currently unselected feature). By the chain rule, $H(S|X)=\sum_{i=1}^{|S|}H(V_i|V_{i-1},\cdots,V_1, X)$ holds.  Under Assumption 2, we have $I(X;S)=H(S)-\sum_{i=1}^{|S|}H(V_i)+\sum_{i=1}^{|S|}I(V_i;X)$.  Since at each iteration $H(S)-\sum_{i=1}^{|S|}H(V_i)$ is the same for all unselected features in $ V\setminus S$ and thus removing them will have no effect on the choice of features, the term $I(X;S)$ in Eq.(\ref{eq7}) is rewritten as $\sum_{i=1}^{|S|}I(V_i;X)\}$  with the pairwise mutual information between a currently unselected feature $X$ and a feature $V_i$ in $S$ without conditioning on other features.

\textbf{Assumption 2.} The selected features in $S$ are conditionally independent given an unselected feature $X\in V\setminus S$ , that is,  $P(S|X) = \prod_{i=1}^{|S|}P(V_i|X)$ where $V_i\in S$.

Under Assumptions 1 and 2,  we reformulate Eq.(\ref{eq7}) as Eq.(\ref{eq8}), i.e., the MRMR objective function. The MRMR approach reduces Eq.(\ref{eq6}) to a linear combination of low-order mutual information terms. To select relevant features with regard to $C$, at each iteration, Eq.(\ref{eq8}) selects the feature $X^*$ that has the maximum relevancy with $C$ and the minimum redundancy with the features in $S$.

\begin{equation}
X^*=\mathop{\arg\max}_{X\in V\setminus S}\{I(X;C)-\sum_{i=1}^{|S|}I(V_i;X)\}.
\label{eq8}
\end{equation}

\subsection{Output of the MRMR approach and the MB of $C$}\label{sec42}

Here we will link the objective function of MRMR to the learning of the MB of C.
From Eq.(\ref{eq8}), we can see that Eq.(\ref{eq8}) only considers three variables, $C$,  the variable in $S$, and the variable outside $S$.  Figure~\ref{fig4-1} shows the three-way variable interactions in a BN.
Using these interactions in Figure~\ref{fig4-1} and the relationship of mutual information and conditional independence discussed in Section~\ref{sec32}, the output of MRMR based on Eq.(\ref{eq8}) is analyzed as follows.


\begin{figure}
\centering
\includegraphics[height=2in,width=2.9in]{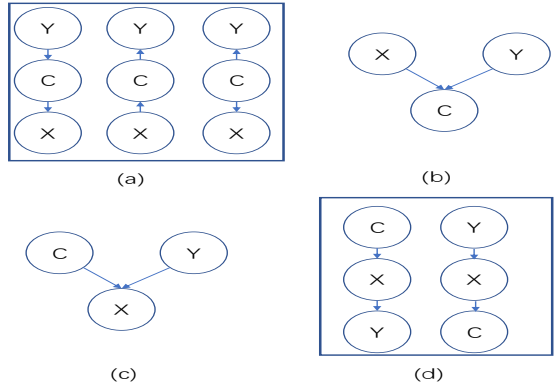}
\caption{Three-way variable interactions in a BN. (a) $X$ and $Y$ are each a parent or a child of $C$; (b) both  $X$ and $Y$ are parents of $C$; (c) $X$ is a common child of $C$ and $Y$ while $Y$ is a spouse of $C$; and (d) $Y$ is a non-child descendant or a non-parent ancestor of $C$.
}
\label{fig4-1}
\end{figure}

\begin{itemize}

\item \textbf{All parents and children of $C$ enter the ouput of MRMR.} In Figures~\ref{fig4-1} (a) and (b), $X$ and $Y$ are each a parent or a child of $C$. No matter whether $X$ or $Y$ enters $S$ or not (i.e., whether $X$ or $Y$ is selected by MRMR or not in the current iteration), by $I(C;(X,Y))=I(C;X)+I(C;Y|X)$, both  $I(C;(X,Y))>I(C;X)$ and $I(C;(X,Y))>I(C;Y)$ hold. In Figure~\ref{fig4-1} (a), $I(X;Y|C)=0$. By $I(X;Y;C)=I(X;C)-I(X;C|Y)=I(Y;C)-I(Y;C|X)=I(X;Y)-I(X;Y|C)$,  we get $I(C;Y)-I(X;Y)$=$I(C;Y)-I(C;Y)+I(C;Y|X)-I(X;Y|C)$=$I(C;Y|X)$. Thus $I(C;Y)-I(X;Y)>0$ holds in Figure~\ref{fig4-1} (a). Similarly, $I(C;X)-I(X;Y)>0$. In Figure~\ref{fig4-1} (b), $I(C;Y)-I(X;Y)$=$I(C;Y)>0$ and $I(C;X)-I(X;Y)>0$. Using Eq.(\ref{eq8}), MRMR selects all parents and children of $C$ in its output.

\item \textbf{Spouses of $C$ do not enter the ouput of MRMR.}  In Figure~\ref{fig4-1} (c), $Y$ is a spouse of $C$ with regard to $X$. By Lemma~\ref{lem3-1}, $I(C;Y)=0$ and $I(X;C)>0$. Thus $I(C;X)>I(C;Y)$ and $X$ will be added to $S$ prior to $Y$. Meanwhile, since $I(C;(X,Y))=I(C;X)+I(C;Y|X)$ and $C\nindep Y|X$ according to Lemma~\ref{lem3-2}, we get that $I(C;(X,Y))>I(C;X)$. This illustrates that a child of $C$ (e.g., $X$)  provides more predictive information of $C$ jointly with $X$'s another parent, i.e., the spouse of $C$ which shares the same child as $C$ (e.g., $Y$) than by itself. Using  Eq.(\ref{eq8}),  assuming that currently $X$ is added to $S$ before $Y$ due to $X$ is a child of $C$, when adding $Y$ to $S$, $I(C;Y)-I(X;Y)<0$ hold. Then using Eq.(\ref{eq8}) MRMR do not include spouses of $C$ in its ouput.

\item \textbf{Non-parent ancestors and non-child descendants of $C$ do not enter the output of MRMR.}  As shown in Figure~\ref{fig4-1} (d), for a non-parent ancestor or a non-child descendant $Y$ of $C$ and a parent or a child $X$ of $C$ on the path from $Y$ to $C$, since $I(C;Y|X)=0$ (i.e., $C\indep Y|X$) and $I(C;X|Y)>0$ (i.e., $X\nindep C|Y$), by $I(C;X)-I(C;X|Y)=I(C;Y)-I(C;Y|X)$, $I(C;X)>I(C;Y)$ holds. Then $X$ will be added to $S$ first. When $X$ is added to $S$, does $\{X,Y\}$ provide more prediction information than $\{X\}$? By $I(C;(X,Y))=I(C;X)+I(C;Y|X)$, we get $I(C;(X,Y))=I(C;X)$, thus given $X$, $Y$ does not provide any information about $C$.
As $X$ is added to $S$, we get $I(C;Y)-I(X;Y)$=$I(C;Y)-I(C;Y)+I(C;Y|X)-I(X;Y|C)$=$-I(X;Y|C)<0$. Then $Y$ cannot be added to the output of MRMR.

In summary, for a non-parent ancestor or a non-child descendant  $Y$,  the parent(s) or child(ren) of $C$ on the path(s) from $Y$ to $C$ always has (have) larger mutual information with $C$, so the parent(s) or child(ren) will be added first, and once added, it is impossible for $Y$ to be added because  $Y$ is independent $C$ given the parent(s) or child(ren).

However, when the sample size is small or noise data samples exist, non-parent ancestors or non-child descendants of $C$ may be added to $S$ prior to parents and children of $C$. The MRMR approach has a strong capability to mitigate problems caused by noise or small-sized data samples, since it calculates pairwise mutual information between features without conditioning on other features.

\end{itemize}

In summary, from the analysis above, we can see that the output of MRMR, i.e., the feature selection based on Eq.(\ref{eq8}) prefers parents and children of $C$.

\subsection{Instantiation of the MRMR approach}\label{sec43}

There are two well-known algorithms which instantiate the MRMR approach, the mRMR (max-Relevance and Min-Redundancy)~\cite{peng2005feature} and FCBF (Fast Correlation Based Filter) algorithms~\cite{yu2004efficient}.  The mRMR algorithm needs to specify the number of PC of a target variable in advance. In fact, we  do not  always have the prior knowledge and thus it is not an easy task to determine a suitable value for the parameter. In contrast, FCBF does not need to specify such a parameter. So in our proposed algorithm (details in Section~\ref{sec5}) we will employ FCBF to learn the PC of each variable.  FCBF has two steps, the forward step (max-relevance) and backward step (min-redundancy).
\begin{itemize}
\item Forward step: FCBF selects a subset of features $S$ that $\forall X\in S$, $I(C;X)>0$, then sorts the features in $S$ by their mutual information with $C$ in descending order.
At this step,  users always use a user-defined threshold $\delta$ ($\delta>0$)  to control the size of  $S$ satisfying $I(C;X)\geq\delta (X\in S)$ instead of $I(C;X)=0$.

\item Backward step: beginning with the first feature $X\in S$, if $\exists Y\in S\setminus\{X\}$ such that  $I(X;Y)>I(X;C)$, then $Y$ is removed from $S$ as a redundant feature to $X$. The FCBF algorithm is terminated until the last feature in $S$ is checked.
\end{itemize}

At the forward step, FCBF only selects features that are relevant to $C$, that is, the candidate PC of $C$. If feature $X$ is a parent or a child of $C$ and $\delta=0$, $I(C;X)>\delta$ should hold. Thus, at the forward step of FCBF, if using  $\delta=0$, all parents and children of $C$ enter $S$. If using  $\delta>0$,  $S$ may include a subset of parents and children of $C$.
At the backward step of FCBF,  for $X\in S$, $Y\in S$, according to the results in Section~\ref{sec42}, we have the following analysis.

First,  if $X$ and $Y$ are each a parent or a child of $C$ (e.g., $X$ and $Y$ in  Figures~\ref{fig4-1} (a) and (b)), $I(X;C)>I(X;Y)$ and $I(Y;C)>I(X;Y)$ hold. Thus both $X$ and $Y$ cannot be used to remove each other from $S$ at the backward step of FCBF.

Second, if  $Y$ is a spouse of $C$ , i.e., both $Y$ and $C$ are parents of $X$ (e.g., $Y$ in  Figure~\ref{fig4-1} (c)), then  $I(C;X)>I(C;Y)$ holds and $X$ is put before $Y$ in $S$.  Since $I(C;Y)<I(X;Y)$, then $Y$ is removed from $S$ at the backward step of FCBF.

Third, if $Y$ is an ancestor or a descendant of $C$ (e.g., $Y$ in  Figure~\ref{fig4-1} (d)),$I(C;X)>I(C;Y)$  and $I(C;Y)<I(X;Y)$ hold.  Since $X$ is put before $Y$ in $S$, $Y$ is removed from $S$.

Thus in theory FCBF prefers parents and children of $C$. At the forward step of FCBF, FCBF can use $\delta$  to control the size of $S$, but  users do not need to specify the number of selected features in advance.

\section{The proposed algorithms}\label{sec5}


Based on the analysis in Section~\ref{sec4} of the MRMR approach and the commonly used algorithms instantiated from MRMR, we can see that the MRMR approach can speed up computing the PC of a variable by  decomposing $I(X;C|S)$  into a linear combination of low-order mutual information terms. In the section, by employing the FCBF algorithm for PC learning, we propose the F2SL approach to local-to-global BN structure learning with the following three steps.

\begin{itemize}
\item Step 1: Learn the PC (parents and children) set of each variable in $V$ using FCBF;

\item Step 2: Construct the DAG skeleton by combining all individual variables¡¯ PC sets;

\item Step 3: Orient edges in the skeleton obtained at Step 2.

\end{itemize}

At Step 2, F2SL constructs the DAG skeleton (i.e., undirected graph)  by combining all the PC sets of individual variables learnt at Step 1 using the symmetry constraint. The symmetry constraint in a BN means that if there is an edge between $V_i$ and $V_j$, the PC set of $V_i$ should contain $V_j$ and the PC set of $V_j$ also should contain $V_i$.
At Step 3, F2SL uses either score functions or conditional independence tests to orient edges in the skeleton obtained at Step 2.

At Step 3, using score functions or independence tests to orient edges,  we instantiate the F2SL approach into two algorithms, F2SL-s and F2SL-c, respectively.

\subsection{The F2SL-s algorithm}

The F2SL-s algorithm is shown in Algorithm 1. It  first learns the PC set of each variable in $V$ using FCBF, then constructs the DAG skeleton using the learnt PC sets. Finally it orients edges in the DAG skeleton using a score function and a hill climbing greedy search algorithm, which is the same as MMHC for edge orientation. Thus the key difference between F2SL-s and MMHC is that F2SL-s uses FCBF to learn a DAG skeleton while MMHC employs the MMPC algorithm~\cite{tsamardinos2003time}.


\begin{algorithm}[t]
\scriptsize
\SetKwData{Left}{left}
  \SetKwData{Up}{up}
  \SetKwFunction{FindCompress}{FindCompress}

  \SetKwRepeat{Repeat}{repeat}{until}
\KwInput{$D$: dataset}

\KwOutput{A (partially) DAG}

Learn PC of each variable in $V$ using FCBF;

Construct the DAG skeleton based on Step 1;

Orient edges using the BDeu score and hill climbing  greedy search;

Output the highest score (partially) DAG.

\caption{\textit{The F2SL-s Algorithm}}
\label{Algorithm 1}
\end{algorithm}

\subsection{F2SL-c algorithm}
Different from F2SL-s, the F2SL-c algorithm (Algorithm 2) employs independence tests for edge orientation at Step 3. Identifying all v-structures in the DAG skeleton is the first and key step to orient edges using independence tests. Thus, for a local structure of three variables, such as $A-C-D$, by Lemma~\ref{lem3-2} of the independence/
dependence relations of a variable (e.g., $A$ and $D$) and its spouses in Section~\ref{sec3}, to determine whether it is a v-structure or not, we need to know the separation set that makes $A$ and $D$ conditionally independent.
For example, Figure~\ref{fig5-1} gives all possible DAGs corresponding to the skeleton  $A-C-D$.
In Figure~\ref{fig5-1} (a), all three DAGs have the same independence/dependence of $A$, $C$, and $D$, i.e., $A\nindep D$ and $A\indep D|C$. The DAG in Figure~\ref{fig5-1} (b) satisfies $A\indep D$ and $A\nindep D|C$, thus it is a v-structure and the separate set of $A$ and $D$ is an empty set.

The separation sets are learned simultaneously with the DAG skeleton using existing MB and PC (Parent-Child) leaning algorithms, however, they have to be learned separately using the MRMR approach based on Eq.(\ref{eq8}).  Since MRMR calculates pairwise mutual information between features without conditioning on other features, it cannot output the separation set of each pair of features. This is a caveat of the fast MRMR approach. To tackle this problem, we propose the FindVstructure algorithm (Algorithm 3) to learn separation sets in order to identifies v-structures in a skeleton.

Then based on the FindVstructure algorithm,  the F2SL-c algorithm is described in Algorithm 2. Before introducing the FindVstructure algorithm,  let us have a look at the different situations/types of separation sets of two variables in a DAG skeleton.
\begin{algorithm}[t]
\scriptsize
\SetKwData{Left}{left}
  \SetKwData{Up}{up}
  \SetKwFunction{FindCompress}{FindCompress}

  \SetKwRepeat{Repeat}{repeat}{until}
\KwInput{$D$: dataset}

\KwOutput{A (partially) DAG}

Learn PC of each variable in $V$ using FCBF;

Construct the DAG skeleton based on Step 1;

Identify all v-structures in the skeleton using FindVstructure;

Orient the remaining edges using the Meek rules;

Output a (partially) DAG.

\caption{\textit{The F2SL-c Algorithm}}
\label{Algorithm 2}
\end{algorithm}

\begin{figure}[t]
\centering
\includegraphics[height=1in,width=3.3in]{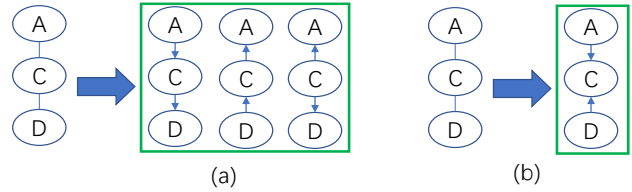}
\caption{Independence/dependence of $A$, $C$, and $D$ and v-structure}
\label{fig5-1}
\end{figure}

\begin{figure}[t]
\centering
\includegraphics[height=1in,width=2.6in]{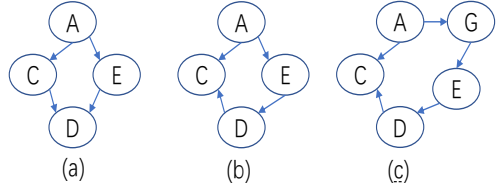}
\caption{Separation set of  $A$ and $D$}
\label{fig5-2}
\end{figure}

\begin{itemize}

\item \textbf{Situation 1: the separation set is an empty set}. In Figure~\ref{fig5-1} (a),  $A$ and $D$ are dependent but become independent given $C$, and $A-C-D$ is a not v-structure. However, in Figure~\ref{fig5-1} (b), $A$ and $D$ are independent but dependent given $C$, and $A-C-D$ is a v-structure. In this case, the separation set of $A$ and $D$ is an empty set.

\item \textbf{Situation 2: the separation set is a subset of the PC set of $A$ or $D$}. In  Figure~\ref{fig5-2} (a), $A$, $C$ and $D$ do not form a v-structure. Thus $A\nindep D|C$ and $A\indep D|C,E$ hold. In  Figure~\ref{fig5-2} (b), $A$, $C$ and $D$ form a v-structure, then $A\indep D|E$ and $A\nindep D|C,E$. In this case, both $E$ and $C$ are the common PC of $A$ and $B$.
In  Figure~\ref{fig5-2} (c), $A\indep D|E$ and $A\indep D|G$. In this case, $E\in PC(D)$ and $G\in PC(A)$. In Figure~\ref{fig5-2} (c), the separation set of $A$ and $D$ is a subset of the PC set of $A$ or $D$.


\end{itemize}

\begin{algorithm}[t]
\scriptsize
\SetKwData{Left}{left}
  \SetKwData{Up}{up}
  \SetKwFunction{FindCompress}{FindCompress}

  \SetKwRepeat{Repeat}{repeat}{until}
\KwInput{Skeleton-DAG}

\KwOutput{Vstructure-DAG}

\For {any local structure $A-C-D$ in Skeleton-DAG }
{
	
   \If {$A\indep D|C$}
          {continue;}

    \If {$A\indep D$ and $A\nindep D|C$}
          {Orient $A-C-D$ as $A\rightarrow C\leftarrow D$;

         continue;
           }



  \If {$\exists S\subseteq PC_{A}$ (or $PC_{D})\setminus \{C\}$ such that $A\indep D|S$ and $A\nindep D|S\cup \{C\}$}
          {Orient $A-C-D$ as $A\rightarrow C\leftarrow D$;
          }

}
\caption{\textit{The FindVstructure Algorithm}}
\label{algorithm 3}
\end{algorithm}

From the above discussions, we see that for a local structure $A-C-D$, a separation set of $A$ and $D$ in a DAG skeleton will be a subset of the PC set of $A$ or $D$. Thus we do not need to search for the separation set for $A$ and $D$ from the union of the PC sets of $A$ and $D$.
With these observations we propose the FindVstructure algorithm (Algorithm 3) to find separation sets and to determine v-structures. For a skeleton $A-C-D$ in a DAG skeleton, instead of searching for  the separation set for $A$ and $D$ from the union of the PC sets of $A$ and $D$,  FindVstructure divides this search into  the  following three steps.

\begin{itemize}

\item Step 1 (Lines 3-5). If $A\indep D|C$ holds,  $A-C-D$ is not a v-structure. Otherwise, go to Step 2.

\item Step 2 (Lines 6-9). If $A\indep D$ and $A\nindep D|C$ hold,  $A-C-D$ is oriented as a v-structure. Otherwise,  go to Step 3.

\item Step 3 (Lines 10-12). Assume $PC(A)$ is the PC set of $A$, Step 3 tests all subsets of $PC(A)$ (or $PC(D)$) excluding $C$. Once finding a subset $S$ that makes both $A\indep D|S$ and $A\nindep D|S\cup\{C\}$ hold, $A-C-D$ is oriented as a v-structure. Otherwise, Step 3 continues until  all subsets within $PC(A)$ are checked.
\end{itemize}

\section{Experiments}\label{sec6}
In this section, we  will systematically evaluate the F2SL-s and F2SL-c algorithms. In Section~\ref{sec60}, we describe the datasets, comparison methods, and evaluation metrics in the experiments. In Section~\ref{sec61}, we  report the results of  F2SL-s and F2SL-c with the five representative local-to-global BN structure learning algorithms, and in Section~\ref{sec62}, we analyze why F2SL-s and F2SL-c are better than the five rivals.

\subsection{Experiment setting}\label{sec60}

To evaluate the F2SL-s and F2SL-c algorithms\footnote{The source codes of F2SL-s/c are available a thttps://github.com/kuiy/CausalLearner}, we use  two groups of data generated from the six benchmark BNs as shown in Table~\ref{tab6-0}\footnote{The data sets are available at $http://pages.mtu.edu/\sim lebrown/supplements/mmhc\_paper/mmhc\_index.html$}. One group includes 10 data sets each with 500 data instances, and the other group also contains 10 data sets each with 1,000 data instances.

All experiments were conducted on a computer with Intel(R) i7-8700, 3.2GHz CPU, and 16GB memory. The significance level for independence tests is set to 0.01. In all Tables in Section~\ref{sec6}, the symbol ``-'' denotes that an algorithm does not produce results when the running time of the algorithm exceeded 48 hours.
We compare the F2SL-s and F2SL-c algorithms against the following five representative local-to-global BN structure learning algorithms:

\begin{itemize}

\item GSBN~\cite{margaritis2000bayesian}. The GSBN algorithm first uses the GSMB algorithm to learning the MB of each variable for constructing the skeleton of a BN, then employs conditional independence tests to orient edges.

\item MMHC~\cite{tsamardinos2006max}. The MMHC algorithm first employs the MMPC algorithm to learning the PC of each variable for constructing the skeleton of a BN, then uses a  score function and a hill climbing greedy search method to orient edges\footnote{The source codes of MMHC are available at http://mensxmachina.org/en/software/probabilistic-graphical-model-toolbox}.

\item  MMHC-c. The MMHC-c algorithm is proposed by us. Compared to MMHC, MMHC-c uses MMPC to learn a DAG skeleton, then it uses conditional independence tests to orient edges.

\item SLL+C/G~\cite{niinimki2012local}. The SLL+C and SLL+G algorithms learn the MB of each variable using a score-based MB learning algorithm, then SLL+C uses conditional independence tests to orient edges while SLL+G employs score functions  to orient edges.\footnote{The source codes of SLL+C/G are available at https://www.cs.helsinki.fi/u/tzniinim/uai2012}

\item GGSL~\cite{gao2017local}. The GGSL algorithm starts to learn a local BN structure of a randomly selected variable using a score-based MB algorithm, then gradually expands the learnt structure until the entire structure is learnt.

\item NOTEARS~\cite{zheng2018dags}. The NOTEARS algorithm learns a global BN structure using continuous optimization and score functions.

\item  DAG-GNN~\cite{yu2019dag}. The DAG-GNN algorithm learns a global BN structure using graph neural  networks.

\end{itemize}

Using the two groups of data sets, in addition to running time (in seconds), we evaluate our algorithms in structure learning quality against the eight rivals from two aspects, structure errors and structure correctness.
The metrics of structural errors (i.e. number of extraneous edges) are described below.

\begin{itemize}
\item SHD (Structural Hamming Distance): the sum of the values of Miss, Extra, and Reverse as follows.
\item Miss:  the number of missing edges in the network structure learnt by the algorithm against the true network structure.
\item Extra:  the number of extra edges in the learnt BN.
\item Reverse: the number of edges with wrong directions according to the true structure.
\end{itemize}

The  metrics of structural correctness (i.e., number of correctly oriented edges)  are Ar\_Precision, Ar\_Recall, and F1 described as follows.

\begin{itemize}

\item Ar\_Recall=$\frac{\#correctly\ predicted\ arrowheads}{\#predicted\ arrowheads}$.

The number of correctly predicted arrowheads in the output  divided by the number of edges in the output of an algorithm.

\item Ar\_Precision=$\frac{\#correctly\ predicted\ arrowheads}{\#true\ arrowheads}$.

The number of correctly predicted arrowheads in the output  divided by the number of true arrowheads in a test DAG.

\item Ar\_F1. $F1 = 2*(Ar\_Precision*Ar\_Recall)/(Ar\_Precision+Ar\_Recall)$. Compared to SHD, $Ar\_F1$ not only considers extraneous edges, but also correct edges.

\end{itemize}

 In the experiments, for an algorithm, we report the average results of these metrics over the ten data sets in each group.

\begin{table}[t]
\scriptsize
\centering
\caption{Summary of benchmark BNs}
\begin{tabular}{|c|c|c|c|c|c|c|c|}
\hline
   Network         & \# & \#   & Max In/Out-  & Min/Max     & Variable \\
     & Vars & Edges  & Degree       & $|$PCset$|$ & Range  \\\hline
Mildew      & 35  & 46      & 3/3          & 1/5         & 3-100  \\\hline
Hailfinder  & 56  & 66      & 4/16         & 1/17        & 2-11    \\\hline
Munin       & 189 & 282     & 3/15         & 1/15        & 1-21    \\\hline
Pigs        & 441 & 592     & 2/39         & 1/41        & 3-3    \\\hline
Link        & 724 & 1,125   & 3/14         & 0/17        & 2-4    \\\hline\
Gene        & 801 & 972     & 4/1          & 0/11        & 3-5    \\
\hline
\end{tabular}
\label{tab6-0}
\end{table}

\begin{table*}[!htbp]
\centering
\centering \caption{Summaries of wrongly learnt edges and time efficiency for the data sets of Group 1 (size=500)}
\scriptsize
\begin{tabular}{ccccccccccccc}
 \hline

Network                      & Algorithm & SHD                    & Reverse      & Miss         & Extra          & Time                \\\hline
\multirow{10}{*}{Mildew}     & GSBN      & 41.40$\pm$0.52             & 5.40$\pm$0.70    & 36.00$\pm$0.82   & 0.00$\pm$0.00      & 0.15$\pm$0.01           \\
                             & MMHC      & 46.20$\pm$1.32             & 9.10$\pm$0.57    & 30.10$\pm$0.88   & 7.00$\pm$0.82      & 0.40$\pm$0.04           \\
                             & MMHC-c    & \textbf{35.30$\pm$1.95}    & 6.00$\pm$0.94    & 26.10$\pm$0.99   & 3.20$\pm$0.63      & 0.24$\pm$0.03           \\
                             & SLL+C     & 41.10$\pm$2.69             & 10.30$\pm$1.16   & 26.90$\pm$1.10   & 3.90$\pm$1.29      & 7,123.30$\pm$9,928.43   \\
                             & SLL+G     & 42.30$\pm$3.30             & 11.70$\pm$2.41   & 26.90$\pm$1.10   & 3.70$\pm$1.25      & 4,537.23$\pm$6,419.42   \\
                             & GGSL      & 45.60$\pm$1.96             & 7.20$\pm$1.32    & 34.50$\pm$1.58   & 3.90$\pm$1.29      & 8,129.13$\pm$6,154.76   \\
                             & NOTEARS   & 56.60$\pm$2.55             & 11.30$\pm$0.82   & 10.60$\pm$1.58   & 34.70$\pm$2.58     & 183.17$\pm$20.28        \\
                             & DAG-GNN   & 41.90$\pm$2.60             & 10.30$\pm$0.48   & 11.20$\pm$1.75   & 20.40$\pm$3.37     & 499.56$\pm$73.88        \\
                             & F2SL-c    & 44.80$\pm$3.58             & 9.90$\pm$2.73    & 21.70$\pm$1.34   & 13.20$\pm$1.62     & 0.23$\pm$0.04           \\
                             & F2SL-s    & 37.40$\pm$1.78             & 12.70$\pm$1.57   & 24.10$\pm$0.57   & 0.60$\pm$0.70      & \textbf{0.14$\pm$0.01}  \\\hline
\multirow{10}{*}{Hailfinder} & GSBN      & \textbf{81.90$\pm$1.20}    & 0.70$\pm$0.48    & 61.00$\pm$0.00   & 20.20$\pm$0.92     & 0.63$\pm$0.00           \\
                             & MMHC      & 107.40$\pm$2.63            & 2.60$\pm$0.84    & 57.50$\pm$0.71   & 47.30$\pm$2.00     & 1.58$\pm$0.07           \\
                             & MMHC-c    & 109.90$\pm$2.56            & 2.30$\pm$0.95    & 56.70$\pm$0.48   & 50.90$\pm$1.97     & 0.66$\pm$0.02           \\
                             & SLL+C     & 99.70$\pm$3.68             & 1.30$\pm$0.48    & 59.70$\pm$1.06   & 38.70$\pm$2.91     & 49.38$\pm$20.99         \\
                             & SLL+G     & 99.70$\pm$3.56             & 2.30$\pm$0.95    & 59.90$\pm$1.20   & 37.50$\pm$2.32     & 31.99$\pm$9.73          \\
                             & GGSL      & 91.50$\pm$10.44            & 1.90$\pm$1.45    & 60.60$\pm$2.63   & 29.00$\pm$11.98    & 369.33$\pm$246.98       \\
                             & NOTEARS   & 198.10$\pm$4.98            & 8.60$\pm$1.51    & 52.60$\pm$1.58   & 136.90$\pm$5.17    & 160.75$\pm$20.36        \\
                             & DAG-GNN   & 104.10$\pm$9.45            & 5.10$\pm$1.10    & 60.30$\pm$1.49   & 38.70$\pm$9.80     & 1,218.35$\pm$81.49       \\
                             & F2SL-c    & 97.30$\pm$2.58             & 0.80$\pm$0.63    & 59.40$\pm$0.97   & 37.10$\pm$2.13     & \textbf{0.19$\pm$0.02}  \\
                             & F2SL-s    & 103.10$\pm$1.45            & 1.00$\pm$0.67    & 58.50$\pm$0.71   & 43.60$\pm$1.35     & 0.26$\pm$0.01           \\\hline
\multirow{10}{*}{Munin}      & GSBN      & \textbf{256.20$\pm$3.22}   & 3.90$\pm$1.52    & 246.80$\pm$3.01  & 5.50$\pm$1.65      & 5.43$\pm$0.05           \\
                             & MMHC      & 322.90$\pm$7.23            & 59.70$\pm$3.68   & 168.10$\pm$4.23  & 95.10$\pm$5.02     & 31.42$\pm$2.35          \\
                             & MMHC-c    & 381.10$\pm$12.78           & 45.40$\pm$3.95   & 178.80$\pm$5.18  & 156.90$\pm$13.08   & 8.11$\pm$0.92           \\
                             & SLL+C     & -             & -            & -            & -              & -                   \\
                             & SLL+G     & -                      & -            & -            & -              & -                   \\
                             & GGSL      & -             & -            & -            & -              & -                   \\
                             & NOTEARS   & 2,795.60$\pm$359.59         & 54.10$\pm$3.51   & 150.20$\pm$7.13  & 2,591.30$\pm$359.87 & 11,113.08$\pm$424.31     \\
                             & DAG-GNN   & 276.50$\pm$5.25            & 4.70$\pm$2.41    & 258.50$\pm$8.46  & 13.30$\pm$4.72     & 10,234.29$\pm$944.72     \\
                             & F2SL-c    & 283.90$\pm$10.40           & 35.00$\pm$5.83   & 165.40$\pm$4.55  & 83.50$\pm$8.71     & \textbf{2.34$\pm$0.09}  \\
                             & F2SL-s    & 320.00$\pm$10.75           & 62.90$\pm$7.72   & 158.70$\pm$2.75  & 98.40$\pm$5.13     & 2.60$\pm$0.11           \\\hline
\multirow{10}{*}{Pigs}       & GSBN      & 297.70$\pm$12.51           & 70.30$\pm$7.54   & 216.60$\pm$6.22  & 10.80$\pm$2.62     & 33.55$\pm$0.60          \\
                             & MMHC      & \textbf{19.80$\pm$8.43}    & 8.10$\pm$4.95    & 0.00$\pm$0.00    & 11.70$\pm$3.80     & 67.49$\pm$0.84          \\
                             & MMHC-c    & 243.50$\pm$7.11            & 105.90$\pm$8.46  & 0.00$\pm$0.00    & 137.60$\pm$4.88    & 36.96$\pm$0.60          \\
                             & SLL+C     & -                      & -            & -            & -              & -                   \\
                             & SLL+G     & -             & -            & -            & -              & -                   \\
                             & GGSL      & -                      & -            & -            & -              & -                   \\
                             & NOTEARS   & 1,842.70$\pm$86.83          & 342.20$\pm$8.46  & 78.50$\pm$7.04   & 1,422.00$\pm$84.03  & 27,374.97$\pm$4,678.84    \\
                             & DAG-GNN   & -                      & -            & -            & -              & -                   \\
                             & F2SL-c    & 61.40$\pm$8.88             & 9.80$\pm$3.36    & 36.80$\pm$3.82   & 14.80$\pm$3.16     & 10.51$\pm$0.16          \\
                             & F2SL-s    & 51.90$\pm$9.37             & 13.80$\pm$4.39   & 23.70$\pm$4.67   & 14.40$\pm$3.20     & \textbf{7.26$\pm$0.17}  \\\hline
\multirow{10}{*}{Link}       & GSBN      & \textbf{1,154.50$\pm$13.87} & 158.10$\pm$8.16  & 911.10$\pm$9.70  & 85.30$\pm$9.75     & 174.41$\pm$8.49         \\
                             & MMHC      & 1,198.60$\pm$26.68          & 130.40$\pm$23.04 & 722.40$\pm$15.90 & 345.80$\pm$22.65   & 187.42$\pm$5.67         \\
                             & MMHC-c    & 1,801.00$\pm$28.03          & 266.70$\pm$18.57 & 591.90$\pm$14.58 & 942.40$\pm$25.22   & 117.52$\pm$11.01        \\
                             & SLL+C     & -             & -            & -            & -              & -                   \\
                             & SLL+G     & -                      & -            & -            & -              & -          \\
                             & GGSL      & -             & -            & -            & -              & -                   \\
                             & NOTEARS   & -                      & -            & -            & -              & -          \\
                             & DAG-GNN   & -             & -            & -            & -              & -                   \\
                             & F2SL-c    & 1,314.20$\pm$72.19          & 233.30$\pm$9.76  & 695.50$\pm$9.59  & 385.40$\pm$70.82   & 67.79$\pm$13.82         \\
                             & F2SL-s    & 1,159.90$\pm$21.57          & 79.50$\pm$9.95   & 724.40$\pm$13.02 & 356.00$\pm$22.90   & \textbf{24.22$\pm$1.02} \\\hline
\multirow{10}{*}{Gene}       & GSBN      & 481.00$\pm$9.50            & 139.40$\pm$9.32  & 328.30$\pm$3.83  & 13.30$\pm$2.67     & 345.63$\pm$15.22        \\
                             & MMHC      & 153.10$\pm$11.62           & 62.40$\pm$9.09   & 46.50$\pm$3.31   & 44.20$\pm$3.55     & 183.39$\pm$1.31         \\
                             & MMHC-c    & 625.30$\pm$29.09           & 264.50$\pm$19.12 & 13.30$\pm$2.98   & 347.50$\pm$11.74   & 75.73$\pm$0.82          \\
                             & SLL+C     & -                      & -            & -            & -              & -                   \\
                             & SLL+G     & -                      & -            & -            & -              & -                   \\
                             & GGSL      & -                      & -            & -            & -              & -                   \\
                             & NOTEARS   & -                      & -            & -            & -              & -          \\
                             & DAG-GNN   & -             & -            & -            & -              & -                   \\
                             & F2SL-c    & \textbf{59.30$\pm$7.96}    & 16.20$\pm$3.91   & 34.90$\pm$4.53   & 8.20$\pm$1.32      & 29.61$\pm$5.37          \\
                             & F2SL-s    & 112.80$\pm$9.76            & 54.60$\pm$7.76   & 49.40$\pm$6.02   & 8.80$\pm$3.12      & \textbf{23.70$\pm$0.64} \\

 \hline
\end{tabular}
\label{tb6-1}
\end{table*}

\begin{table*}[!htbp]
\centering
 \caption{Summaries of wrongly learnt edges and time efficiency for the data sets of Group 2 (size=1,000)}
\scriptsize
\begin{tabular}{ccccccccccccc}
 \hline
Network                      & Algorithm & SHD                    & Reverse      & Miss        & Extra          & Time                \\\hline
\multirow{10}{*}{Mildew}     & GSBN      & 41.00$\pm$1.05             & 4.00$\pm$0.47    & 36.90$\pm$0.88  & 0.10$\pm$0.32      & 0.22$\pm$0.00           \\
                             & MMHC      & 39.40$\pm$2.99             & 8.60$\pm$1.26    & 25.50$\pm$1.18  & 5.30$\pm$1.34      & 0.49$\pm$0.03           \\
                             & MMHC-c    & 33.60$\pm$1.58             & 7.20$\pm$0.79    & 21.40$\pm$0.84  & 5.00$\pm$1.25      & 0.28$\pm$0.01           \\
                             & SLL+C     & 33.80$\pm$0.84             & 7.20$\pm$0.45    & 22.60$\pm$0.55  & 4.00$\pm$0.00      & 6,867.33$\pm$1,970.90   \\
                             & SLL+G     & 39.00$\pm$1.73             & 12.20$\pm$1.92   & 22.80$\pm$0.45  & 4.00$\pm$0.00      & 5,553.74$\pm$1,640.63   \\
                             & GGSL      & -                      & -            & -           & -              & -                   \\
                             & NOTEARS   & 53.10$\pm$1.45             & 11.30$\pm$0.67   & 10.40$\pm$0.97  & 31.40$\pm$2.07     & 185.71$\pm$9.99         \\
                             & DAG-GNN   & 53.90$\pm$8.10             & 10.60$\pm$0.70   & 11.30$\pm$2.16  & 32.00$\pm$7.27     & 947.51$\pm$91.00        \\
                             & F2SL-c    & \textbf{24.80$\pm$2.66}    & 5.60$\pm$1.43    & 15.00$\pm$0.82  & 4.20$\pm$1.14      & \textbf{0.13$\pm$0.01}  \\
                             & F2SL-s    & 32.40$\pm$2.12             & 10.60$\pm$1.71   & 20.60$\pm$0.52  & 1.20$\pm$0.79      & 0.15$\pm$0.01           \\\hline
\multirow{10}{*}{Hailfinder} & GSBN      & \textbf{80.80$\pm$0.79}    & 0.10$\pm$0.32    & 60.00$\pm$0.00  & 20.70$\pm$0.67     & 0.85$\pm$0.02           \\
                             & MMHC      & 107.40$\pm$0.97            & 1.30$\pm$0.82    & 57.20$\pm$0.63  & 48.90$\pm$0.57     & 1.70$\pm$0.05           \\
                             & MMHC-c    & 112.00$\pm$2.83            & 2.60$\pm$0.70    & 56.10$\pm$0.74  & 53.30$\pm$2.63     & 0.78$\pm$0.07           \\
                             & SLL+C     & 96.00$\pm$1.05             & 1.30$\pm$0.67    & 57.80$\pm$0.63  & 36.90$\pm$0.88     & 36.06$\pm$12.59         \\
                             & SLL+G     & 99.70$\pm$3.56             & 2.30$\pm$0.95    & 59.90$\pm$1.20  & 37.50$\pm$2.32     & 31.99$\pm$9.73          \\
                             & GGSL      & 93.00$\pm$2.21             & 2.30$\pm$0.67    & 57.50$\pm$0.53  & 33.20$\pm$2.04     & 234.68$\pm$146.90       \\
                             & NOTEARS   & 184.70$\pm$4.19            & 7.70$\pm$0.82    & 54.00$\pm$0.67  & 123.00$\pm$4.40    & 160.65$\pm$21.82        \\
                             & DAG-GNN   & 121.00$\pm$7.94            & 6.70$\pm$0.82    & 57.20$\pm$1.40  & 57.10$\pm$8.20     & 2,167.98$\pm$231.95      \\
                             & F2SL-c    & 95.80$\pm$2.62             & 0.40$\pm$0.52    & 59.80$\pm$0.79  & 35.60$\pm$2.37     & \textbf{0.24$\pm$0.02}  \\
                             & F2SL-s    & 100.70$\pm$1.34            & 0.40$\pm$0.70    & 58.10$\pm$0.32  & 42.20$\pm$0.92     & 0.27$\pm$0.01           \\\hline
\multirow{10}{*}{Munin}      & GSBN      & \textbf{246.90$\pm$3.21}   & 6.50$\pm$1.27    & 232.50$\pm$2.42 & 7.90$\pm$1.85      & 6.57$\pm$0.06           \\
                             & MMHC      & 302.50$\pm$4.62            & 66.30$\pm$3.89   & 151.30$\pm$2.45 & 84.90$\pm$3.60     & 45.35$\pm$3.51          \\
                             & MMHC-c    & 479.60$\pm$14.21           & 67.40$\pm$2.95   & 141.30$\pm$3.27 & 270.90$\pm$13.69   & 14.39$\pm$0.87          \\
                             & SLL+C     & -             & -            & -           & -              & -                   \\
                             & SLL+G     & -                      & -            & -           & -              & -                   \\
                             & GGSL      & -             & -            & -           & -              & -                   \\
                             & NOTEARS   & 2,482.40$\pm$272.50         & 53.90$\pm$5.57   & 145.20$\pm$6.49 & 2,283.30$\pm$272.13 & 11,362.21$\pm$576.29     \\
                             & DAG-GNN   & 291.20$\pm$31.08           & 12.90$\pm$2.23   & 215.30$\pm$8.15 & 63.00$\pm$36.24    & 12,552.36$\pm$685.19     \\
                             & F2SL-c    & 252.70$\pm$3.40            & 24.30$\pm$4.55   & 160.40$\pm$1.71 & 68.00$\pm$3.13     & \textbf{2.98$\pm$0.09}  \\
                             & F2SL-s    & 309.10$\pm$9.85            & 65.30$\pm$5.98   & 150.30$\pm$3.77 & 93.50$\pm$3.54     & 3.05$\pm$0.10           \\\hline
\multirow{10}{*}{Pigs}       & GSBN      & 280.70$\pm$8.54            & 65.30$\pm$8.92   & 208.80$\pm$7.74 & 6.60$\pm$0.97      & 37.94$\pm$0.53          \\
                             & MMHC      & \textbf{9.00$\pm$4.50}     & 5.00$\pm$3.62    & 0.00$\pm$0.00   & 4.00$\pm$2.05      & 74.09$\pm$1.00          \\
                             & MMHC-c    & 207.70$\pm$11.56           & 63.90$\pm$6.79   & 0.00$\pm$0.00   & 143.80$\pm$7.10    & 44.70$\pm$1.04          \\
                             & SLL+C     & -                      & -            & -           & -              & -                   \\
                             & SLL+G     & -             & -            & -           & -              & -                   \\
                             & GGSL      & -                      & -            & -           & -              & -                   \\
                             & NOTEARS   & -             & -            & -           & -              & -                   \\
                             & DAG-GNN   & -                      & -            & -           & -              & -                   \\
                             & F2SL-c    & 29.40$\pm$8.24             & 5.20$\pm$2.74    & 17.70$\pm$5.19  & 6.50$\pm$2.51      & 13.90$\pm$0.25          \\
                             & F2SL-s    & 23.90$\pm$7.09             & 5.40$\pm$2.95    & 11.20$\pm$4.05  & 7.30$\pm$2.50      & \textbf{8.93$\pm$0.03}  \\\hline
\multirow{10}{*}{Link}       & GSBN      & \textbf{1,095.50$\pm$11.40} & 163.80$\pm$5.83  & 828.80$\pm$9.58 & 102.90$\pm$7.05    & 178.85$\pm$4.69         \\
                             & MMHC      & 1,120.70$\pm$32.66          & 138.20$\pm$23.43 & 650.00$\pm$6.45 & 332.50$\pm$14.81   & 200.64$\pm$3.76         \\
                             & MMHC-c    & 1,766.90$\pm$29.21          & 222.70$\pm$15.90 & 546.20$\pm$5.39 & 998.00$\pm$16.72   & 129.53$\pm$5.22         \\
                             & SLL+C     & -             & -            & -           & -              & -                   \\
                             & SLL+G     & -                      & -            & -           & -              & -          \\
                             & GGSL      & -             & -            & -           & -              & -                   \\
                             & NOTEARS   & -             & -            & -           & -              & -          \\
                             & DAG-GNN   & -                      & -            & -           & -              & -                   \\
                             & F2SL-c    & 1,237.10$\pm$21.00          & 196.20$\pm$15.50 & 666.60$\pm$6.40 & 374.30$\pm$12.54   & 65.66$\pm$2.99          \\
                             & F2SL-s    & 1,149.20$\pm$16.59          & 95.50$\pm$11.61  & 663.70$\pm$6.53 & 390.00$\pm$12.44   & \textbf{32.77$\pm$1.68} \\\hline
\multirow{10}{*}{Gene}       & GSBN      & 471.60$\pm$8.81            & 133.50$\pm$7.47  & 324.40$\pm$2.63 & 13.70$\pm$3.16     & 348.09$\pm$15.09        \\
                             & MMHC      & 113.00$\pm$12.02           & 58.80$\pm$9.96   & 28.60$\pm$3.34  & 25.60$\pm$4.17     & 190.79$\pm$0.96         \\
                             & MMHC-c    & 586.20$\pm$21.19           & 226.70$\pm$15.11 & 3.00$\pm$0.82   & 356.50$\pm$11.94   & 91.08$\pm$13.82         \\
                             & SLL+C     & -                      & -            & -           & -              & -                   \\
                             & SLL+G     & -                      & -            & -           & -              & -                   \\
                             & GGSL      & -                      & -            & -           & -              & -                   \\
                             & NOTEARS   & -                      & -            & -           & -              & -                   \\
                             & DAG-GNN   & -                      & -            & -           & -              & -                   \\
                             & F2SL-c    & \textbf{50.70$\pm$5.23}    & 11.50$\pm$2.84   & 33.40$\pm$2.95  & 5.80$\pm$0.92      & 45.51$\pm$4.76          \\
                             & F2SL-s    & 94.70$\pm$10.07            & 52.50$\pm$10.38  & 35.00$\pm$2.58  & 7.20$\pm$1.62      & \textbf{29.40$\pm$0.27} \\
 \hline
\end{tabular}
\label{tb6-2}
\end{table*}

\begin{table*}[!htbp]
\centering
\centering \caption{Comparison of correctly learnt edge directions}
\scriptsize
\begin{tabular}{cc||ccc|cccccccc}
\hline
                             &           & Size=500           &           &           & Size=1,000         &           &           \\
Network                      & Algorithm & F1                 & Precision & Recall    & F1                 & Precision & Recall    \\\hline
\multirow{10}{*}{Mildew}     & GSBN      & 0.16$\pm$0.02          & 0.46$\pm$0.05 & 0.10$\pm$0.01 & 0.18$\pm$0.03          & 0.55$\pm$0.07 & 0.11$\pm$0.02 \\
                             & MMHC      & 0.20$\pm$0.03          & 0.30$\pm$0.03 & 0.15$\pm$0.02 & 0.33$\pm$0.05          & 0.46$\pm$0.07 & 0.26$\pm$0.04 \\
                             & MMHC-c    & 0.40$\pm$0.05          & 0.60$\pm$0.06 & 0.30$\pm$0.04 & 0.46$\pm$0.03          & 0.59$\pm$0.03 & 0.38$\pm$0.03 \\
                             & SLL+C     & 0.26$\pm$0.05          & 0.38$\pm$0.07 & 0.19$\pm$0.03 & 0.44$\pm$0.02          & 0.59$\pm$0.02 & 0.35$\pm$0.02 \\
                             & SLL+G     & 0.21$\pm$0.08          & 0.32$\pm$0.12 & 0.16$\pm$0.06 & 0.30$\pm$0.05          & 0.40$\pm$0.07 & 0.24$\pm$0.04 \\
                             & GGSL      & 0.14$\pm$0.05          & 0.28$\pm$0.09 & 0.09$\pm$0.04 & -                  & -         & -         \\
                             & NOTEARS   & 0.42$\pm$0.02          & 0.34$\pm$0.02 & 0.52$\pm$0.03 & 0.43$\pm$0.01          & 0.36$\pm$0.01 & 0.53$\pm$0.02 \\
                             & DAG-GNN   & \textbf{0.48$\pm$0.03} & 0.44$\pm$0.02 & 0.53$\pm$0.04 & 0.43$\pm$0.05          & 0.36$\pm$0.05 & 0.52$\pm$0.05 \\
                             & F2SL-c    & 0.34$\pm$0.07          & 0.38$\pm$0.08 & 0.31$\pm$0.07 & \textbf{0.63$\pm$0.05} & 0.72$\pm$0.06 & 0.55$\pm$0.04 \\
                             & F2SL-s    & 0.27$\pm$0.05          & 0.41$\pm$0.07 & 0.20$\pm$0.04 & 0.41$\pm$0.05          & 0.56$\pm$0.07 & 0.32$\pm$0.04 \\\hline
\multirow{10}{*}{Hailfinder} & GSBN      & 0.09$\pm$0.01          & 0.17$\pm$0.02 & 0.07$\pm$0.01 & \textbf{0.13$\pm$0.01} & 0.22$\pm$0.01 & 0.09$\pm$0.00 \\
                             & MMHC      & 0.10$\pm$0.02          & 0.11$\pm$0.02 & 0.09$\pm$0.02 & 0.12$\pm$0.02          & 0.13$\pm$0.02 & 0.11$\pm$0.02 \\
                             & MMHC-c    & 0.11$\pm$0.02          & 0.12$\pm$0.02 & 0.11$\pm$0.02 & 0.11$\pm$0.01          & 0.12$\pm$0.01 & 0.11$\pm$0.01 \\
                             & SLL+C     & 0.09$\pm$0.02          & 0.11$\pm$0.03 & 0.08$\pm$0.02 & 0.12$\pm$0.01          & 0.15$\pm$0.01 & 0.10$\pm$0.01 \\
                             & SLL+G     & 0.07$\pm$0.03          & 0.09$\pm$0.04 & 0.06$\pm$0.03 & 0.11$\pm$0.03          & 0.15$\pm$0.04 & 0.09$\pm$0.03 \\
                             & GGSL      & 0.07$\pm$0.04          & 0.11$\pm$0.07 & 0.05$\pm$0.03 & 0.12$\pm$0.02          & 0.15$\pm$0.02 & 0.09$\pm$0.02 \\
                             & NOTEARS   & 0.04$\pm$0.01          & 0.03$\pm$0.01 & 0.07$\pm$0.01 & 0.04$\pm$0.00          & 0.03$\pm$0.00 & 0.07$\pm$0.01 \\
                             & DAG-GNN   & 0.01$\pm$0.01          & 0.01$\pm$0.01 & 0.01$\pm$0.01 & 0.03$\pm$0.02          & 0.03$\pm$0.02 & 0.03$\pm$0.02 \\
                             & F2SL-c    & \textbf{0.11$\pm$0.02} & 0.13$\pm$0.02 & 0.09$\pm$0.01 & 0.11$\pm$0.02          & 0.14$\pm$0.02 & 0.09$\pm$0.01 \\
                             & F2SL-s    & \textbf{0.11$\pm$0.01} & 0.13$\pm$0.01 & 0.10$\pm$0.01 & \textbf{0.13$\pm$0.01} & 0.15$\pm$0.01 & 0.11$\pm$0.01 \\\hline
\multirow{10}{*}{Munin}      & GSBN      & 0.19$\pm$0.01          & 0.77$\pm$0.05 & 0.11$\pm$0.01 & 0.25$\pm$0.01          & 0.75$\pm$0.03 & 0.15$\pm$0.01 \\
                             & MMHC      & 0.22$\pm$0.01          & 0.26$\pm$0.02 & 0.19$\pm$0.01 & 0.26$\pm$0.01          & 0.30$\pm$0.01 & 0.23$\pm$0.01 \\
                             & MMHC-c    & 0.21$\pm$0.01          & 0.22$\pm$0.02 & 0.20$\pm$0.01 & 0.21$\pm$0.01          & 0.18$\pm$0.01 & 0.26$\pm$0.01 \\
                             & SLL+C     & -         & -         & -         & -         & -         & -         \\
                             & SLL+G     & -                  & -         & -         & -                  & -         & -         \\
                             & GGSL      & -         & -         & -         & -         & -         & -         \\
                             & NOTEARS   & 0.05$\pm$0.01          & 0.03$\pm$0.01 & 0.28$\pm$0.03 & 0.06$\pm$0.01          & 0.03$\pm$0.00 & 0.29$\pm$0.01 \\
                             & DAG-GNN   & 0.12$\pm$0.04          & 0.51$\pm$0.06 & 0.07$\pm$0.02 & 0.26$\pm$0.02          & 0.43$\pm$0.07 & 0.19$\pm$0.03 \\
                             & F2SL-c    & \textbf{0.34$\pm$0.02} & 0.41$\pm$0.03 & 0.29$\pm$0.02 & \textbf{0.41$\pm$0.01} & 0.51$\pm$0.02 & 0.35$\pm$0.01 \\
                             & F2SL-s    & 0.24$\pm$0.03          & 0.27$\pm$0.04 & 0.21$\pm$0.03 & 0.26$\pm$0.03          & 0.29$\pm$0.03 & 0.24$\pm$0.03 \\\hline
\multirow{10}{*}{Pigs}       & GSBN      & 0.62$\pm$0.02          & 0.79$\pm$0.02 & 0.52$\pm$0.02 & 0.65$\pm$0.02          & 0.82$\pm$0.02 & 0.54$\pm$0.01 \\
                             & MMHC      & \textbf{0.98$\pm$0.01} & 0.97$\pm$0.01 & 0.99$\pm$0.01 & \textbf{0.99$\pm$0.01} & 0.98$\pm$0.01 & 0.99$\pm$0.01 \\
                             & MMHC-c    & 0.74$\pm$0.01          & 0.67$\pm$0.01 & 0.82$\pm$0.01 & 0.80$\pm$0.01          & 0.72$\pm$0.01 & 0.89$\pm$0.01 \\
                             & SLL+C     & -                  & -         & -         & -                  & -         & -         \\
                             & SLL+G     & -         & -         & -         & -         & -         & -         \\
                             & GGSL      & -                  & -         & -         & -                  & -         & -         \\
                             & NOTEARS   & 0.14$\pm$0.01          & 0.09$\pm$0.01 & 0.29$\pm$0.01 & -                  & -         & -         \\
                             & DAG-GNN   & -                  & -         & -         & -                  & -         & -         \\
                             & F2SL-c    & 0.94$\pm$0.01          & 0.96$\pm$0.01 & 0.92$\pm$0.01 & 0.97$\pm$0.01          & 0.98$\pm$0.01 & 0.96$\pm$0.01 \\
                             & F2SL-s    & 0.94$\pm$0.01          & 0.95$\pm$0.01 & 0.94$\pm$0.01 & 0.98$\pm$0.01          & 0.98$\pm$0.01 & 0.97$\pm$0.01 \\\hline
\multirow{10}{*}{Link}       & GSBN      & 0.08$\pm$0.01          & 0.19$\pm$0.03 & 0.05$\pm$0.01 & 0.17$\pm$0.01          & 0.33$\pm$0.01 & 0.12$\pm$0.01 \\
                             & MMHC      & 0.29$\pm$0.02          & 0.36$\pm$0.03 & 0.24$\pm$0.02 & 0.35$\pm$0.02          & 0.42$\pm$0.03 & 0.30$\pm$0.02 \\
                             & MMHC-c    & 0.20$\pm$0.01          & 0.18$\pm$0.01 & 0.24$\pm$0.01 & 0.26$\pm$0.01          & 0.23$\pm$0.01 & 0.32$\pm$0.02 \\
                             & SLL+C     & -                  & -         & -         & -                  & -         & -         \\
                             & SLL+G     & -                  & -         & -         & -                  & -         & -         \\
                             & GGSL      & -                  & -         & -         & -                  & -         & -         \\
                             & NOTEARS   & -                  & -         & -         & -                  & -         & -         \\
                             & DAG-GNN   & -                  & -         & -         & -                  & -         & -         \\
                             & F2SL-c    & 0.20$\pm$0.01          & 0.24$\pm$0.02 & 0.17$\pm$0.01 & 0.27$\pm$0.02          & 0.31$\pm$0.02 & 0.23$\pm$0.01 \\
                             & F2SL-s    & \textbf{0.34$\pm$0.01} & 0.42$\pm$0.02 & 0.29$\pm$0.01 & \textbf{0.37$\pm$0.01} & 0.43$\pm$0.01 & 0.33$\pm$0.01 \\\hline
\multirow{10}{*}{Gene}       & GSBN      & 0.62$\pm$0.01          & 0.77$\pm$0.01 & 0.52$\pm$0.01 & 0.63$\pm$0.01          & 0.78$\pm$0.01 & 0.53$\pm$0.01 \\
                             & MMHC      & 0.89$\pm$0.01          & 0.89$\pm$0.01 & 0.89$\pm$0.01 & 0.91$\pm$0.01          & 0.91$\pm$0.01 & 0.91$\pm$0.01 \\
                             & MMHC-c    & 0.61$\pm$0.02          & 0.53$\pm$0.02 & 0.71$\pm$0.02 & 0.65$\pm$0.01          & 0.56$\pm$0.01 & 0.76$\pm$0.02 \\
                             & SLL+C     & -                  & -         & -         & -                  & -         & -         \\
                             & SLL+G     & -                  & -         & -         & -                  & -         & -         \\
                             & GGSL      & -                  & -         & -         & -                  & -         & -         \\
                             & NOTEARS   & -         & -         & -         & -         & -         & -         \\
                             & DAG-GNN   & -                  & -         & -         & -                  & -         & -         \\
                             & F2SL-c    & \textbf{0.96$\pm$0.01} & 0.97$\pm$0.00 & 0.95$\pm$0.01 & \textbf{0.97$\pm$0.00} & 0.98$\pm$0.00 & 0.95$\pm$0.01 \\
                             & F2SL-s    & 0.91$\pm$0.01          & 0.93$\pm$0.01 & 0.89$\pm$0.01 & 0.92$\pm$0.01          & 0.94$\pm$0.01 & 0.91$\pm$0.01 \\
\hline
\end{tabular}
\label{tb6-3}
\end{table*}

\begin{figure}[!htbp]
\centering
\includegraphics[width=3.0in, height=2in, angle=0]{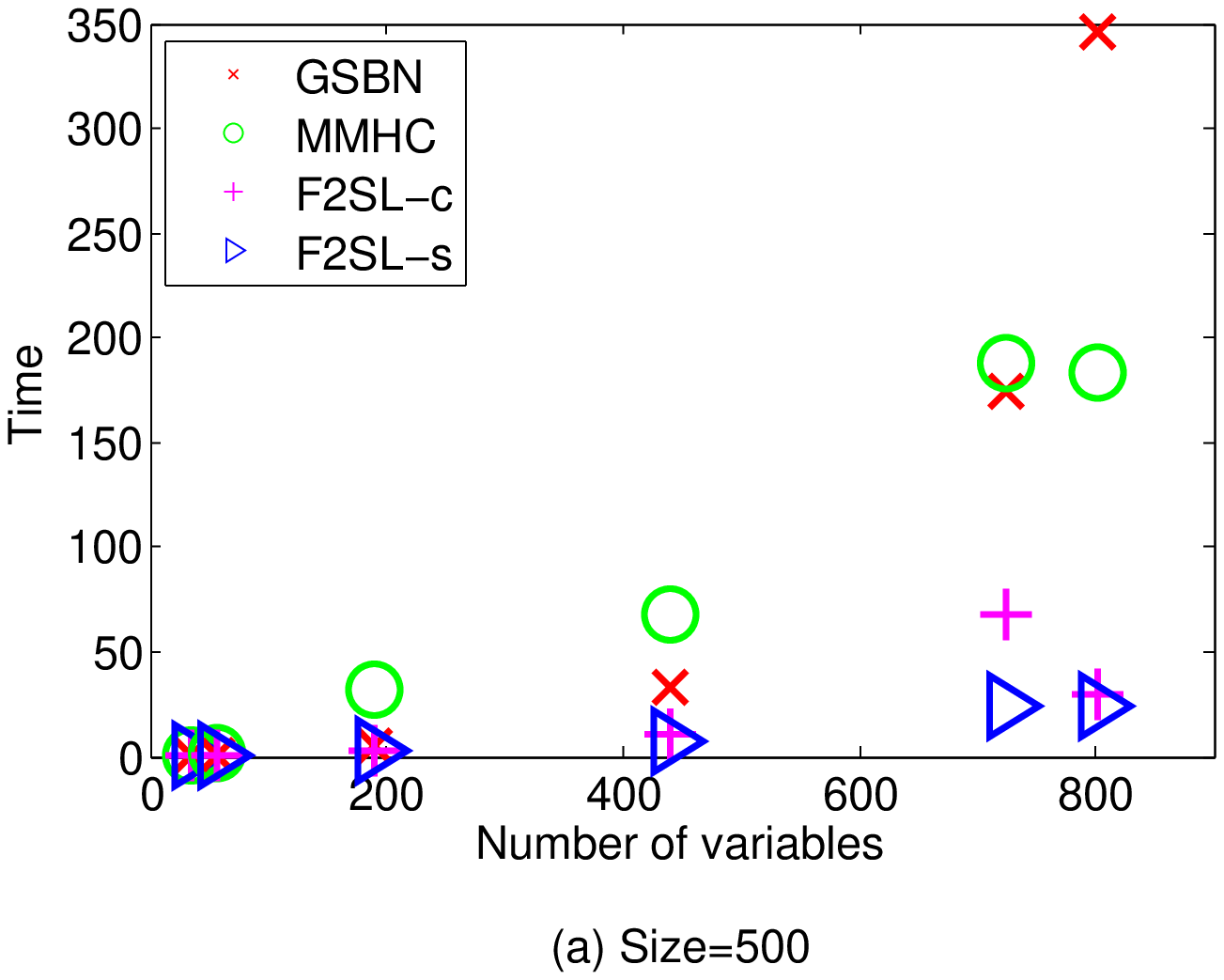}
\caption{Running time
with varying variables using sample size 500 (each point denotes running time of an algorithm on a BN in Table~\ref{tab6-0})}
\label{fig6-01}
\end{figure}

\begin{figure}[!htbp]
\centering
\includegraphics[width=3.0in, height=2in, angle=0]{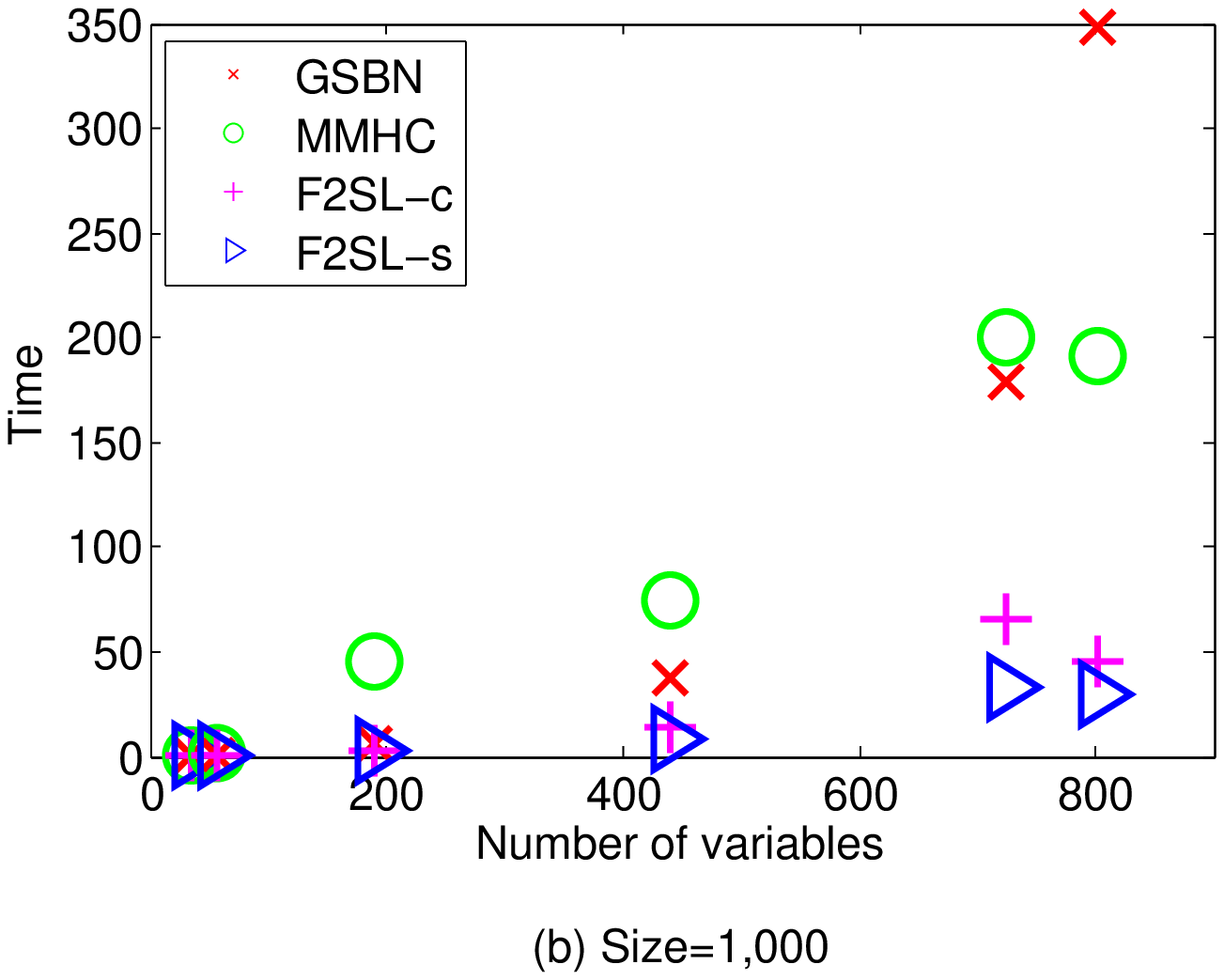}
\caption{Running time with varying variables using sample size 1000 (each point denotes running time of an algorithm on a BN in Table~\ref{tab6-0})}
\label{fig6-02}
\end{figure}

\subsection{Results of Bayesian network structure learning}\label{sec61}

In this section, we will report the experimental results of F2SL-c and F2SL-s vs. the eight rivals in terms of time efficiency and quality of learned structures, respectively.

\textbf{1. Time efficiency}. The last columns in Tables~\ref{tb6-1} and~\ref{tb6-2} show that F2SL-c and F2SL-s are significantly faster than GSBN, MMHC, SLL+C, SLL+G, MMHC-C, GGSL, NOTEARS, and DAG-GNN in both groups of datasets.  GGSL, SLL+C,  NOTEARS, DAG-GNN, and SLL+G are not scalable to a large size of a BN network. Based on the normalized running time results shown in Table~\ref{tb6-4}, F2SL-s is up to 15, 15, 50800, 32408, 58065 times faster than GSBN, MMHC,  SLL+C, SLL+G, and GGSL, respectively. Figures~\ref{fig6-01} and~\ref{fig6-02} show that the running time of GSBN and MMHC increases with the size of BNs networks while F2SL-c and F2SL-s  are much more scalable.


\textbf{2. Structure errors.}
Tables~\ref{tb6-1} to~\ref{tb6-2} show the BN quality by SHD of F2SL-c and F2SL-s against the eight rivals.

\textbf{(1) F2SL-c and F2SL-s against GSBN.} We can see that for both groups of datasets (with 500 and 1000 samples respectively), F2SL-c and F2SL-s  are inferior to GSBN on the Hailfinder, Munin, and Link networks, since the values of both Extra and Reverse of GSBN are much smaller than those of F2SL-c and F2SL-s.
GSBN employs the GSMB algorithm to find the MB of each feature for constructing the DAG skeleton. Since GSMB uses the entire feature subset currently selected for computing Eq.(\ref{eq6}), the number of data samples required by GSMB is exponential to the size of the MB. In Hailfinder, Munin, and Link networks, for both groups of datasets, the size of the MB of each feature selected by GSMB is much small. As a result, GSBN misses many true edges (i.e. has higher Miss values), but smaller Extra and Reverse of GSBN than F2SL-c and F2SL-s make it overall SHD lower than F2SL-c and F2SL-s. 

\textbf{(2)
F2SL-c and F2SL-s against MMHC and MMHC-c.}
Using both groups of datasets, we have the following observations.
F2SL algorithms learned a DAG skeleton better than MMHC. Both F2SL-c and MMHC-c have the same orientation process and the difference in their performance lies in learning skeletons. The BN structures learnt by MMHC-c have much more extra edges than those learnt by F2SL-c, but have slightly less missing edges than the BN structures learnt by F2SL-c. This indicates that F2SL finds more accurate skeletons than MMHC.
Edge orientation in F2SL-c (by a conditional independence test) is better than edge orientation in MMHC (by a score function). The edge orientation by a scoring function greedily deletes edges during edge orientation, and the deletion reducers the performance of BN learning since this edge deletion procedure not only deletes wrong edges, but may also remove correct edges. MMHC and F2SL-s employ the same orientation strategy. The skeleton learnt by F2SL is better than that learnt by MMHC as we discussed above, but after edge orientations, the performance of MMHC and F2SL-s are similar and the reduced performance of F2SL-s is due to edge deletion. To confirm this, we compare F2SL-c, which does not delete edges in the orientation process, with MMHC. On the Mildew, Hailfinder, Munin, and Gene networks, F2SL-c achieves better SHD than MMHC, and this confirms that deleting edges hurts the overall performance.
Pigs is a special data set for MMHC which achieve 0 missing edges (100\% recall rate as shown in Table~\ref{tb6-5} in Section~\ref{sec62}). Such an exceptional performance does not repeat in other data sets.

\textbf{(3)
F2SL-c and F2SL-s against SLL+C/G.}
Both SLL+C and SLL+G only produce the results on the Mildew and Hailfinder networks due to expensively computational costs. We can see that the SHD values of both SLL+C and SLL+G are very close to those of F2SL-c and F2SL-s, but SLL+C and SLL+G are very computationally expensive.

\textbf{(4)
F2SL-c and F2SL-s against NOTEARS and DAG-GNN.}
From Tables~\ref{tb6-1} to~\ref{tb6-2}, we can see that NOTEARS and DAG-GNN  are computationally expensive and are not scalable to large and/or high dimensional data sets, thus they do not produce the results on the Pigs, Link and Gene networks.
NOTEARS and DAG-GNN often achieve a smaller number of missing edges but a much larger number of extra edges than F2SL-c and F2SL-s.  In small BN networks, the SHD values of DAG-GNN are competitive with those of F2SL-c and F2SL-s.

\textbf{3. Structural correctness.}
Table~\ref{tb6-3} reports the BN structure quality by Ar\_Recall, Ar\_Precision, and Ar\_F1 of F2SL-c and F2SL-s and their rivals. Although on the Hailfinder, Munin, and Link networks, GSBN achieves less structure errors than F2SL-c and F2SL-s in terms of $SHD$,  it is inferior to F2SL-c/s on Ar\_F1 with both groups of data as shown in Table~\ref{tb6-3}. As discussed above, GSMB employed by GSBN finds much samller sized MBs than FCBF used by F2SL-c and F2SL-s, and thus GSMB has missed more true edges than F2SL-c and F2SL-s, and GSBN achieves much lower value of Ar\_Recall than F2SL-c and F2SL-s.

F2SL-c and F2SL-s are better than its eight rivals on all networks except for the Pigs network. F2SL-c and F2SL-s learn more correct edges than its eight rivals. From Tables~\ref{tb6-1} and~\ref{tb6-2}, F2SL-s and MMHC are very competitive in terms of Miss, thus they are comparable on Ar\_F1 as shown in Table~\ref{tb6-3}. The Pigs network is a very special for MMHC which achieves 100\% recall (as shown in Table~\ref{tb6-5}).

From Table~\ref{tb6-3}, we can see that F2SL-c and F2SL-s achieves larger Recall than  NOTEARS and DAG-GNN (except for the Mildew network using 500 samples), although NOTEARS and DAG-GNN get smaller Miss than F2SL-c and F2SL-s (as shown in Tables~\ref{tb6-1} and~\ref{tb6-2}). This is because NOTEARS and DAG-GNN have large orientation errors. Overall, F2SL-s and F2SL-s achieve better Ar\_F1 than NOTEARS and DAG-GNN.

\textbf{4. Simultaneous comparison of time efficiency and structure quality.}
In BN structure learning,  an algorithm may be chosen to sacrifice structure learning quality for computational efficiency, while another algorithm may be chosen to sacrifice time efficiency for structure learning quality.
Therefore it is interesting to compare two algorithms in terms of both time efficiency and quality of structure learning at the same time.

For this comparison, we normalize SHD, Ar\_F1, and running time reported in Tables~\ref{tb6-1} to~\ref{tb6-3}. Normalized SHD (or Ar\_F1 or running time) is the value of SHD  (or Ar\_F1 or running time) of an algorithm for a particular sample size and network divided by the SHD (or Ar\_F1 or running time) of F2SL-s on the same sample size and network.
The normalized results are shown in Table~\ref{tb6-4}.  A normalized SHD or running time greater than one implies  that the algorithm is worse or slower than F2SL-s on the same learning task, while a normalized Ar\_F1 smaller than one implies that the algorithm is worse than F2SL-s on the same learning task.

Using the normalized results in Table~\ref{tb6-4},  we explore the trade-off between structure quality (i.e. SHD or  Ar\_F1) and  time efficiency of each algorithm.  Figure~\ref{fig6-11} shows the logarithm of the normalized time versus the logarithm of the normalized SHD for the two groups of datasets with sample sizes 500 and 1000 respectively, and Figure~\ref{fig6-14} illustrates the logarithm of the normalized time vs. the normalized Ar\_F1 for the two groups of datasets with sample sizes 500 and 1000 respectively.
Each point in Figures~\ref{fig6-11} and~\ref{fig6-14} denotes the performance in terms of the two metrics of a given algorithm on learning one of the six BN networks.  Since we are using normalized measures with respect to the measures of F2SL-s,  the  performance measures of F2SL-s always fall on point $(1,1)$  and thus are not indicated in the figures.

We can see that in Figure~\ref{fig6-11}, there are no algorithms (except for F2SL-c) fall in the grey area, indicating that no algorithms outperform F2SL-s in terms of both running time and SHD. In terms of SHD, F2SL-s  and F2SL-c are very competitive with the other algorithms in most cases while no algorithms are faster than  F2SL-s  and F2SL-c.

Figure~\ref{fig6-14} illustrates that no algorithms (except for F2SL-c) fall in the grey area, indicating that no algorithms are better than F2SL-s  in terms of both running time and  Ar\_F1, except for F2SL-c. GGSL  is better than F2SL-s only in one case in the left figure of Figure~\ref{fig6-14} and  SLL+C is superior to F2SL-s in one case in the right figure of Figure~\ref{fig6-14}. Except for the two cases, no other algorithms outperform  F2SL-s  and F2SL-c.


In summary,  in terms of time efficiency, F2SL-c and F2SL-s are significantly faster than GSMB, MMHC, GGSL, SLL+C, SLL+G, NOTEARS and DAG-GNN. Both of them achieve competitive  performance against their eight rivals in terms of the metrics of structural errors. In terms of the metrics of structural correctness, F2SL-c and F2SL-s are better than their five rivals. In the following, we will analyze why F2SL-c and F2SL-s are better than these rivals.




\begin{figure*}[t]
\centering
       \begin{tabular}{ccccc}
      \subfigure{\includegraphics[width=3.0in, height=2.3in]{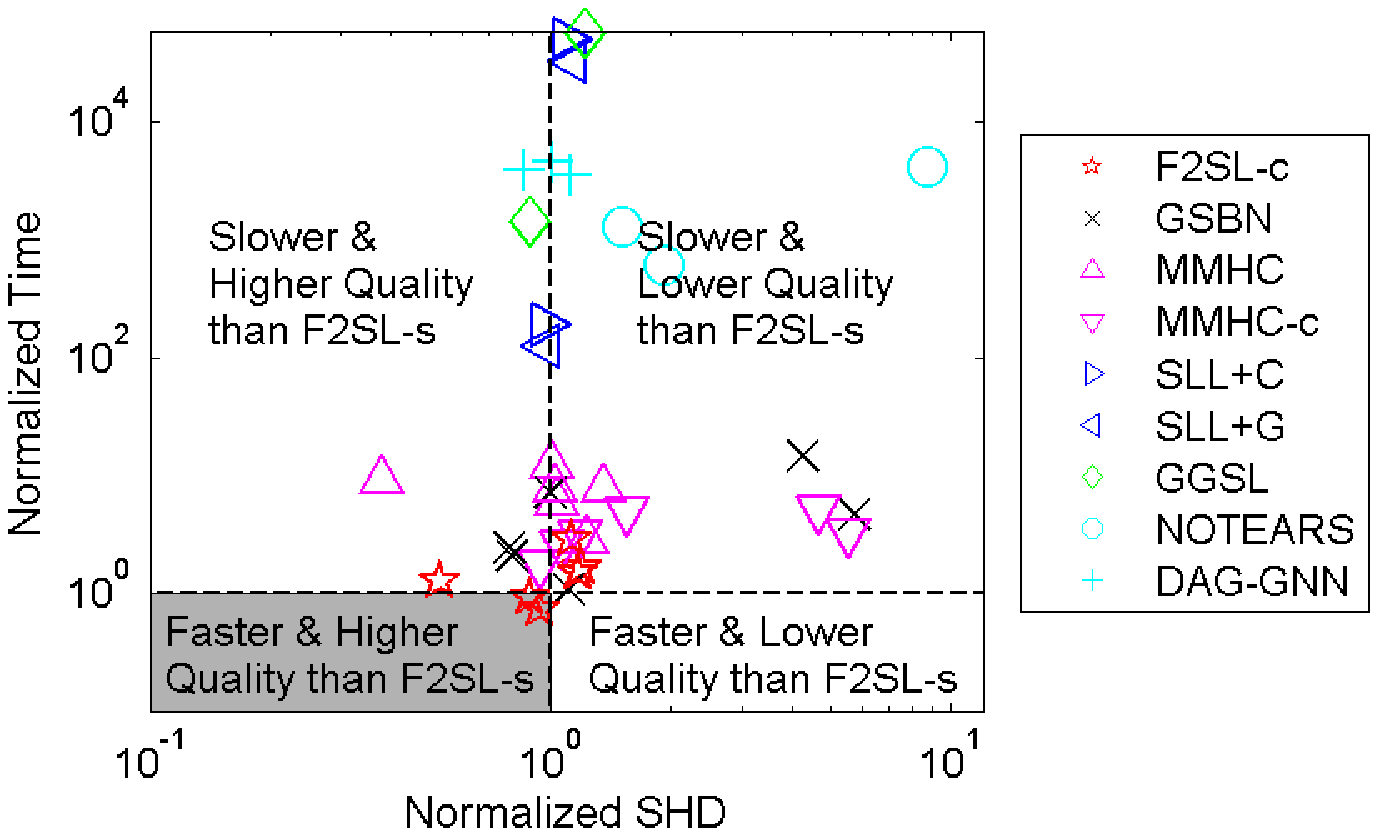}}\vspace{-0.4cm}&
      \subfigure{\includegraphics[width=3.0in, height=2.3in]{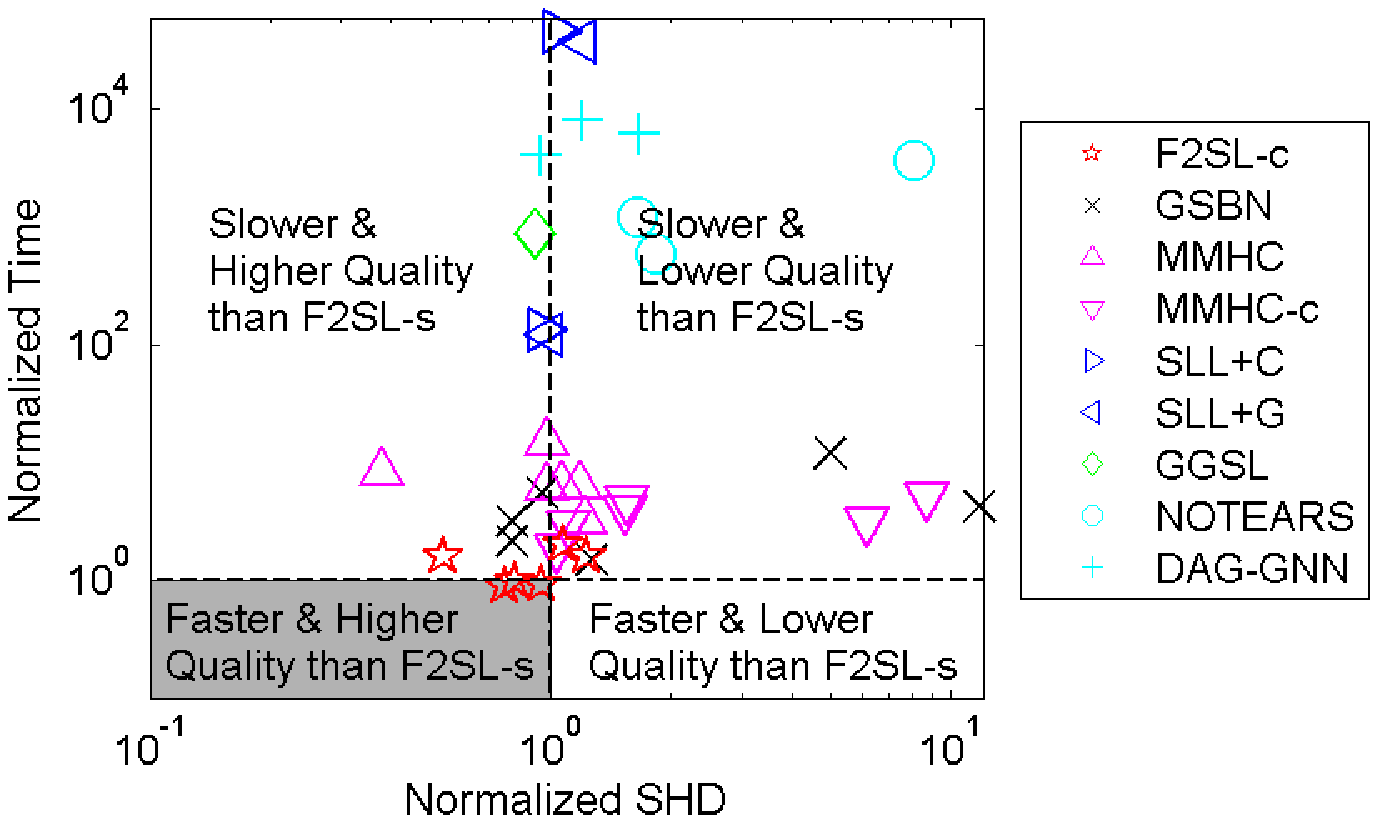}}\vspace{-0.4cm}&
       \end{tabular}
 \caption{Normalized Time vs. Normalized SHD with 500 (left) and 1000 (right) data samples}
 \label{fig6-11}
\end{figure*}

\begin{table*}[t]
\centering
\centering \caption{Comparison of structure learning methods on different groups of data sets in terms of normalized metrics ($\downarrow$ means the smaller, the better while $\uparrow$ denotes that the bigger, the better)}
\scriptsize
\begin{tabular}{c||cccccc|ccccccc}
\hline
        & Size=500  &            &       &      &      &       & Size=1,000 &            &       &       &      &       \\
Network & Mildew    & Hailfinder & Munin & Pigs & Link & Gene  & Mildew     & Hailfinder & Munin & Pigs  & Link & Gene  \\\hline
        & \multicolumn{12}{c}{\textbf{Normalized Time $\downarrow$}}                                                                 \\
F2SL-s  & 1.00      & 1.00       & 1.00  & 1.00 & 1.00 & 1.00  & 1.00       & 1.00       & 1.00  & 1.00  & 1.00 & 1.00  \\
F2SL-c  & 1.64      & 0.73       & 0.90  & 1.45 & 2.80 & 1.25  & 0.87       & 0.89       & 0.98  & 1.56  & 2.00 & 1.55  \\
GSBN    & 1.07      & 2.42       & 2.09  & 4.62 & 7.20 & 14.58 & 1.47       & 3.15       & 2.15  & 4.25  & 5.46 & 11.84 \\
MMHC    & 2.86      & 6.08       & 12.08 & 9.30 & 7.74 & 7.74  & 3.27       & 6.30       & 14.87 & 8.30  & 6.12 & 6.49  \\
MMHC-c  & 1.71      & 2.54       & 3.12     & 5.09     & 4.85 & 3.20  & 1.87       & 2.89       & 4.72     & 5.01  & 3.95 & 3.10  \\
SLL+C   & 50,880.71 & 189.92     & -     & -    & -    & -     & 45,782.20  & 133.56     & -     & -     & -    & -     \\
SLL+G   & 32,408.79 & 123.04     & -     & -    & -    & -     & 37,024.93  & 120.41     & -     & -     & -    & -     \\
GGSL    & 58,065.21 & 1,420.50   & -     & -    & -    & -     & -          & 869.19     & -     & -     & -    & -     \\
NOTEARS & 1,308.36  & 618.27     & 4,274.26 & 3,770.66 & -    & -     & 1,238.07   & 595.00     & 3,725.31 & -     & -    & -     \\
DAG-GNN & 3,568.29  & 4,685.96   & 3,936.27 & -        & -    & -     & 6,316.73   & 8,029.56   & 4,115.53 & -     & -    & -     \\\hline
        & \multicolumn{12}{c}{\textbf{Normalized SHD $\downarrow$}}                                                                  \\
F2SL-s  & 1.00      & 1.00       & 1.00  & 1.00 & 1.00 & 1.00  & 1.00       & 1.00       & 1.00  & 1.00  & 1.00 & 1.00  \\
F2SL-c  & 1.20      & 0.94       & 0.89  & 1.18 & 1.13 & 0.53  & 0.77       & 0.95       & 0.82  & 1.23  & 1.08 & 0.54  \\
GSBN    & 1.11      & 0.79       & 0.80  & 5.74 & 1.00 & 4.26  & 1.27       & 0.80       & 0.80  & 11.74 & 0.95 & 4.98  \\
MMHC    & 1.24      & 1.04       & 1.01  & 0.38 & 1.03 & 1.36  & 1.22       & 1.07       & 0.98  & 0.38  & 0.98 & 1.19  \\
MMHC-c  & 0.94      & 1.07       & 1.19     & 4.69     & 1.55 & 5.54  & 1.04       & 1.11       & 1.55     & 8.69  & 1.54 & 6.19  \\
SLL+C   & 1.10      & 0.97       & -     & -    & -    & -     & 1.04       & 0.95       & -     & -     & -    & -     \\
SLL+G   & 1.13      & 0.97       & -     & -    & -    & -     & 1.20       & 0.99       & -     & -     & -    & -     \\
GGSL    & 1.22      & 0.89       & -     & -    & -    & -     & -          & 0.92       & -     & -     & -    & -     \\
NOTEARS & 1.51      & 1.92       & 8.74     & 35.50    & -    & -     & 1.64       & 1.83       & 8.03     & -     & -    & -     \\
DAG-GNN & 1.12      & 1.01       & 0.86     & -        & -    & -     & 1.66       & 1.20       & 0.94     & -     & -    & -     \\\hline
        & \multicolumn{12}{c}{\textbf{Normalized Ar\_F1 $\uparrow$}}                                                               \\
F2SL-s  & 1.00      & 1.00       & 1.00  & 1.00 & 1.00 & 1.00  & 1.00       & 1.00       & 1.00  & 1.00  & 1.00 & 1.00  \\
F2SL-c  & 1.26      & 1.00       & 1.42  & 1.00 & 0.59 & 1.05  & 1.54       & 0.85       & 1.58  & 0.99  & 0.73 & 1.05  \\
GSBN    & 0.59      & 0.82       & 0.79  & 0.66 & 0.24 & 0.68  & 0.44       & 1.00       & 0.96  & 0.66  & 0.46 & 0.68  \\
MMHC    & 0.74      & 0.91       & 0.92  & 1.04 & 0.85 & 0.98  & 0.80       & 0.92       & 1.00  & 1.01  & 0.95 & 0.99  \\
MMHC-c  & 1.48      & 1.00       & 0.88     & 0.79     & 0.59 & 0.67  & 1.12       & 0.85       & 0.81     & 0.82  & 0.70 & 0.71  \\
SLL+C   & 0.96      & 0.82       & -     & -    & -    & -     & 1.07       & 0.92       & -     & -     & -    & -     \\
SLL+G   & 0.78      & 0.64       & -     & -    & -    & -     & 0.73       & 0.85       & -     & -     & -    & -     \\
GGSL    & 0.52      & 0.64       & -     & -    & -    & -     & -          & 0.92       & -     & -     & -    & -     \\
NOTEARS & 1.56      & 0.36       & 0.21     & 0.15     & -    & -     & 1.05       & 0.31       & 0.23     & -     & -    & -     \\
DAG-GNN & 1.78      & 0.09       & 0.50     & -        & -    & -     & 1.05       & 0.23       & 1.00     & -     & -    & -     \\\hline
\end{tabular}
\label{tb6-4}
\end{table*}

\begin{figure*}[t]
\centering
       \begin{tabular}{ccccc}
      \subfigure{\includegraphics[width=3.0in, height=2.3in]{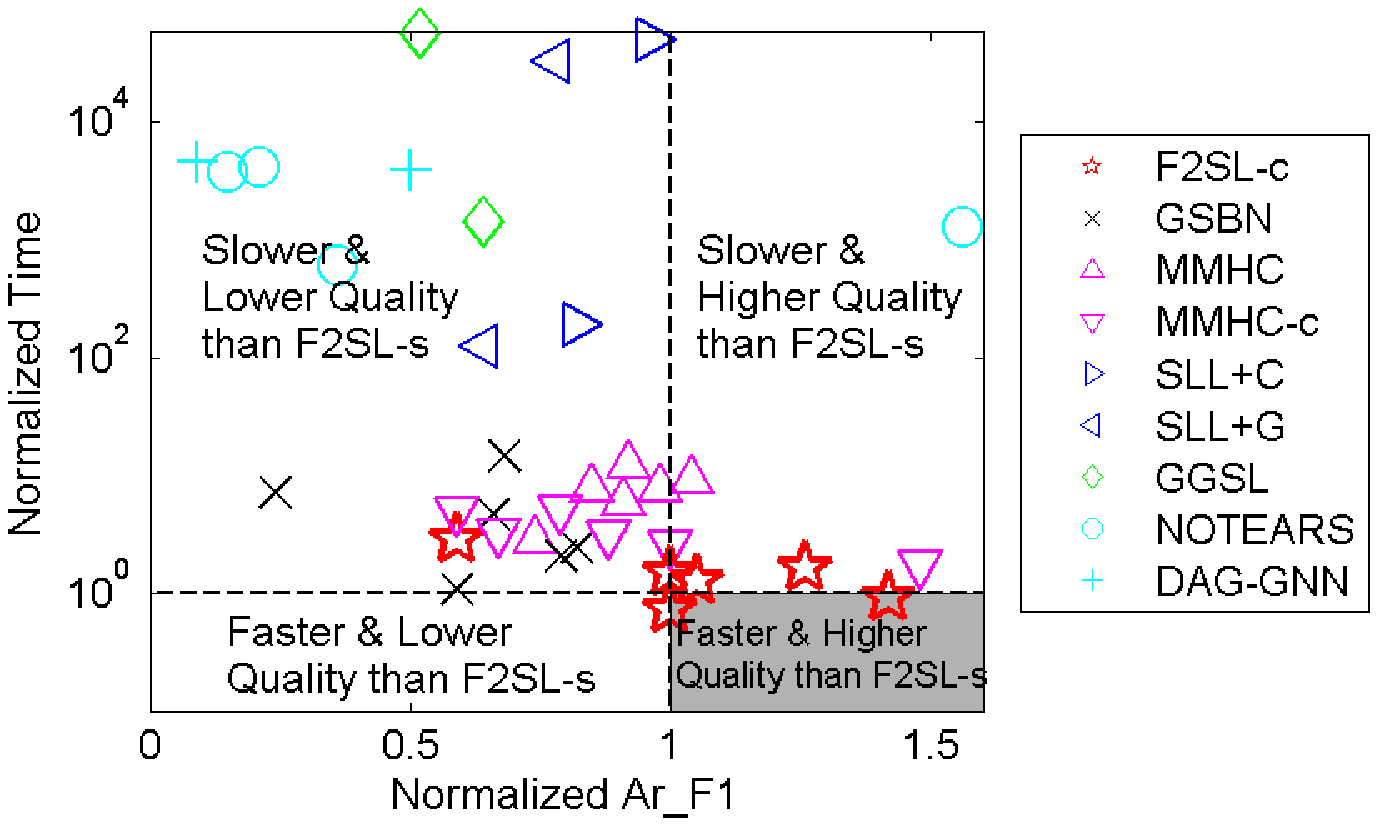}}\vspace{-0.4cm}&
      \subfigure{\includegraphics[width=3.0in, height=2.3in]{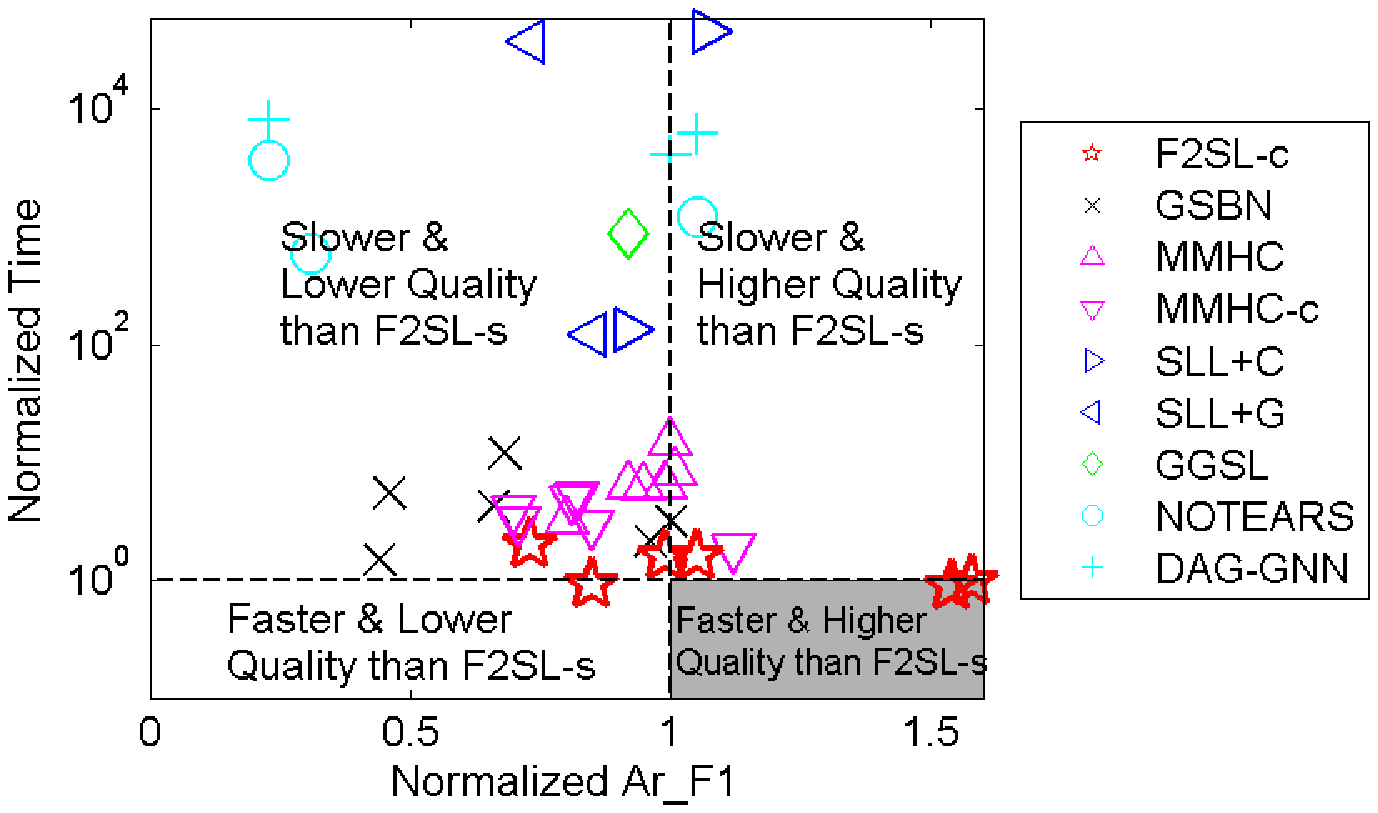}}\vspace{-0.4cm}&
       \end{tabular}
 \caption{Normalized Time vs. Normalized Ar\_F1 with 500 (left) and 1000 (right) data samples}
 \label{fig6-14}
\end{figure*}

\subsection{Why the proposed algorithms are better?}\label{sec62}

The efficiency and quality of MB or PC learning for each feature are the key to local-to-global BN learning algorithms. Among F2SL-c, F2SL-s and their five rivals,  GSBN employs the GSMB algorithm for learning the MB (not the PC) of each feature to constructing  skeletons,  whereas MMHC, SLL+C, SLL+G, GGSL, F2SL-c, and F2SL-s learn the PC set of each feature for constructing the skeletons.
GSBN employs a different method to learn the skeleton of a BN from MMHC, SLL+C, SLL+G, GGSL, F2SL-c, and F2SL-s.
In this section, we evaluate the performance of the PC learning algorithms used by F2SL-c, F2SL-s, MMHC, SLL+C, SLL+G, and GGSL to find out why F2SL-c and F2SL-s are better. For the evaluation of the performance of the PC learning algorithms, we use precision and recall as described below (and for time efficiency evaluation, we use running time).

\begin{itemize}

\item Precision. The number of true positives in the output of a PC learning algorithm (i.e. the number of variables in the output belonging to the true  PC of a target variable) divided by the total number of variables in the output.

\item Recall. The number of true positives in the output divided by the size of the true  PC set of a variable.

\item F1. $F1 = 2*(precision*recall)/(precision+recall)$.

\end{itemize}


\subsubsection{FCBF versus standard PC learning algorithm.}

As described in Section~\ref{sec5}, F2SL-c and F2SL-s employ the FCBF algorithm to learn the PC of each variable for skeleton construction,  and MMHC uses the MMPC algorithm~\cite{tsamardinos2003time} to find PC of each variable.
SLL+C/G  and GGSL employ SLL-PC and $S^2$TMB-PC to learn the PC of each variable in a data set for  skeleton construction.  
In this section, in addition to MMPC, SLL-PC, and $S^2$TMB-PC, we do a comprehensive comparison of FCBF with another two widely used PC learning algorithms, HITON-PC~\cite{aliferis2003hiton} and  PC-simple~\cite{li2015practical} to investigate why F2SL-c and F2SL-s are better than MMHC, SLL+C, SLL+G,  and GGSL.

Table~\ref{tb6-5} shows the PC learning quality and running time of the PC learning algorithms. Table~\ref{tb6-6} gives the normalized running time  and F1.
Normalized F1 or running time is the value of F1 or running time of an algorithm for a particular sample size and network divided by the F1 or running time of F2SL-s on the same sample size and network.
In Table~\ref{tb6-6},  a normalized running time greater than one indicates that an algorithm is slower than FCBF, while a normalized F1 smaller than one implies that an algorithm is worse than FCBF.

From Tables~\ref{tb6-5} and~\ref{tb6-6}, we can see that FCBF is significantly faster than the other PC methods on both groups of data sets. This explains why F2SL-c and F2SL-s  are much more efficient than the other five BN structure learning algorithms.  SLL-PC and $S^2$TMB-PC are so computationally expensive, and this explains why  SLL+C, SLL+G, and GGSL are not scalable to large-sized networks.

In PC learning quality,  in terms of F1, FCBF is better than MMPC, HITON-PC, SLL-PC and $S^2$TMB-PC on almost all networks. FCBF achieves almost the same F1 value as PC-simple on the Hilfinder, Pigs, and Gene networks, but it is significantly better than PC-simple on the remaining networks.
In terms of recall,  FCBF is very competitive with its rivals while it achieves better precision than these rivals, except for SLL-PC and $S^2$TMB-PC on the Mildew network. However, SLL-PC and $S^2$TMB-PC are significantly worse than  FCBF in terms of recall. Thus on average F2SL-c and F2SL-s are better than SLL+C, SLL+G, and GGSL. On the Hilfinder network,  FCBF, SLL-PC,  and $S^2$TMB-PC are very comparable, thus F2SL-c and F2SL-s  are competitive with SLL+C, SLL+G, and GGSL.

For FCBF and MMPC, MMPC calculates the high order mutual information between $X$ and $C$ conditioning on a subset of the already selected features $S$. In the worst case, MMPC needs to explore all possible subsets of $S$. When the size of PC sets becomes large, MMPC will be very computationally expensive and require large data samples for reliable independence tests. A large size of PC sets and/or small-sized data samples will make MMPC impractical in many real-world applications. Just as we discussed in Section 4, FCBF can deal with small and/or high dimensional datasets, since FCBF computes the pairwise mutual information between $X$ and $C$ without conditioning on other features.

Table~\ref{tb6-5} has shown that FCBF achieves higher precision rate than and almost the same recall rate as MMPC although FCBF only uses pairwise mutual information to learn an approximate PC set of a variable. When the size of a PC set is large (e.g. in Munin) or a BN network has a large number of variables (e.g. in Gene), FCBF is much better than MMPC on both recall and precision measures. In  Tables~\ref{tb6-1} and~\ref{tb6-2} in Section~\ref{sec61}, we observed that a higher recall achieved by MMPC does not transfer to smaller missing edges in BN structures learnt by MMHC in comparison with F2SL-c. FCBF achieves higher precision than MMPC. This means that the output of MMPC contains more false PCs than FCBF and hence the BN structures learnt by MMHC and MMHC-c should have more extra edges than those learnt by F2SL-c.  But MMHC achieves a competitive number of extra edges with F2SL-c (except for the Gene network) due to edge deletion at the edge orientation step, while  the BN structures learnt by MMHC-c have more extra edges and less missing edges than those  learnt by F2SL-c on all networks except for the Mildew network due to non-edge deletion.
In summary, for local-to-global structure learning, regardless of which method, score-based or constraint-based method, is used for edge orientations, it is crucial to find a correct DAG skeleton.

\begin{table*}[t]
\centering
\centering \caption{Comparison of  FCBF with four PC learning algorithms}
\scriptsize
\begin{tabular}{cc||cccc|ccccccc}
\hline
                           &               & Size=500           &           &           &                    & Size=1,000         &           &           &                    \\
Network                    & Algorithm     & F1                 & Precision & Recall    & Time               & F1                 & Precision & Recall    & Time               \\\hline
\multirow{6}{*}{Mildew}    & MMPC          & 0.53$\pm$0.01          & 0.77$\pm$0.02 & 0.46$\pm$0.01 & 0.01$\pm$0.00          & 0.70$\pm$0.02          & 0.75$\pm$0.02 & 0.72$\pm$0.02 & 0.02$\pm$0.00          \\
                           & HITONPC       & 0.51$\pm$0.02          & 0.51$\pm$0.04 & 0.61$\pm$0.03 & 0.01$\pm$0.00          & 0.51$\pm$0.03          & 0.46$\pm$0.03 & 0.70$\pm$0.03 & 0.02$\pm$0.00          \\
                           & PC-simple     & 0.27$\pm$0.01          & 0.17$\pm$0.01 & 0.86$\pm$0.02 & 0.61$\pm$0.04          & 0.32$\pm$0.01          & 0.21$\pm$0.01 & 0.87$\pm$0.03 & 0.43$\pm$0.01          \\
                           & SLL-PC        & 0.53$\pm$0.03          & 0.71$\pm$0.04 & 0.47$\pm$0.03 & 283.56$\pm$189.87      & \textbf{-}         & -         & -         & \textbf{-}         \\
                           & S$^{2}$TMB-PC & 0.54$\pm$0.03          & 0.71$\pm$0.03 & 0.48$\pm$0.03 & 241.16$\pm$142.72      & -                  & -         & -         & \textbf{-}         \\
                           & FCBF          & \textbf{0.59$\pm$0.03} & 0.65$\pm$0.03 & 0.58$\pm$0.03 & \textbf{0.00$\pm$0.00} & \textbf{0.75$\pm$0.02} & 0.83$\pm$0.02 & 0.73$\pm$0.02 & \textbf{0.00$\pm$0.00} \\\hline
\multirow{6}{*}{Hilfinder} & MMPC          & 0.12$\pm$0.01          & 0.14$\pm$0.01 & 0.15$\pm$0.01 & 0.03$\pm$0.00          & 0.13$\pm$0.01          & 0.14$\pm$0.01 & 0.15$\pm$0.01 & 0.04$\pm$0.00          \\
                           & HITONPC       & \textbf{0.13$\pm$0.01} & 0.14$\pm$0.01 & 0.15$\pm$0.01 & 0.04$\pm$0.00          & 0.13$\pm$0.01          & 0.14$\pm$0.01 & 0.15$\pm$0.01 & 0.05$\pm$0.00          \\
                           & PC-simple     & \textbf{0.13$\pm$0.01} & 0.14$\pm$0.01 & 0.14$\pm$0.01 & 0.03$\pm$0.00          & 0.13$\pm$0.00          & 0.14$\pm$0.00 & 0.15$\pm$0.01 & 0.04$\pm$0.00          \\
                           & SLL-PC        & 0.12$\pm$0.01          & 0.13$\pm$0.02 & 0.12$\pm$0.02 & 7.01$\pm$2.40          & \textbf{0.14$\pm$0.01} & 0.15$\pm$0.01 & 0.14$\pm$0.01 & 5.15$\pm$1.51          \\
                           & S$^{2}$TMB-PC & 0.12$\pm$0.01          & 0.14$\pm$0.01 & 0.13$\pm$0.01 & 6.96$\pm$3.63          & \textbf{0.14$\pm$0.01} & 0.15$\pm$0.01 & 0.15$\pm$0.01 & 4.18$\pm$2.48          \\
                           & FCBF          & \textbf{0.13$\pm$0.01} & 0.16$\pm$0.01 & 0.13$\pm$0.01 & \textbf{0.00$\pm$0.00} & 0.13$\pm$0.01          & 0.17$\pm$0.01 & 0.13$\pm$0.01 & \textbf{0.00$\pm$0.00} \\\hline
\multirow{6}{*}{Munin}     & MMPC          & 0.13$\pm$0.01          & 0.11$\pm$0.02 & 0.43$\pm$0.02 & 0.66$\pm$0.10          & 0.20$\pm$0.01          & 0.17$\pm$0.01 & 0.53$\pm$0.02 & 0.96$\pm$0.12          \\
                           & HITONPC       & 0.13$\pm$0.01          & 0.11$\pm$0.02 & 0.43$\pm$0.02 & 9.99$\pm$1.35          & 0.21$\pm$0.01          & 0.18$\pm$0.01 & 0.53$\pm$0.02 & 5.50$\pm$0.46          \\
                           & PC-simple     & -                  & -         & -         & -                  & -                  & -         & -         & -                  \\
                           & SLL-PC        & \textbf{-}         & -         & -         & \textbf{-}         & \textbf{-}         & -         & -         & -                  \\
                           & S$^{2}$TMB-PC & -                  & -         & -         & \textbf{-}         & -                  & -         & -         & \textbf{-}         \\
                           & FCBF          & \textbf{0.36$\pm$0.01} & 0.34$\pm$0.01 & 0.47$\pm$0.02 & \textbf{0.01$\pm$0.00} & \textbf{0.41$\pm$0.01} & 0.40$\pm$0.01 & 0.48$\pm$0.01 & \textbf{0.01$\pm$0.00} \\\hline
\multirow{6}{*}{Pigs}      & MMPC          & 0.91$\pm$0.00          & 0.87$\pm$0.01 & 1.00$\pm$0.00 & 0.10$\pm$0.00          & 0.91$\pm$0.00          & 0.86$\pm$0.01 & 1.00$\pm$0.00 & 0.15$\pm$0.00          \\
                           & HITONPC       & 0.91$\pm$0.00          & 0.87$\pm$0.01 & 1.00$\pm$0.00 & 0.13$\pm$0.00          & 0.91$\pm$0.00          & 0.86$\pm$0.01 & 1.00$\pm$0.00 & 0.17$\pm$0.00          \\
                           & PC-simple     & \textbf{0.99$\pm$0.00} & 0.98$\pm$0.00 & 1.00$\pm$0.00 & 0.18$\pm$0.01          & \textbf{0.99$\pm$0.00} & 0.98$\pm$0.00 & 1.00$\pm$0.00 & 0.27$\pm$0.01          \\
                           & SLL-PC        & -                  & -         & -         & -                  & -                  & -         & -         & -                  \\
                           & S$^{2}$TMB-PC & -                  & -         & -         & -                  & -                  & -         & -         & -                  \\
                           & FCBF          & 0.96$\pm$0.01          & 0.96$\pm$0.00 & 0.96$\pm$0.01 & \textbf{0.01$\pm$0.00} & 0.98$\pm$0.01          & 0.98$\pm$0.01 & 0.98$\pm$0.01 & \textbf{0.02$\pm$0.00} \\\hline
\multirow{6}{*}{Link}      & MMPC          & 0.37$\pm$0.01          & 0.39$\pm$0.01 & 0.44$\pm$0.01 & 0.42$\pm$0.47          & 0.40$\pm$0.01          & 0.41$\pm$0.01 & 0.47$\pm$0.01 & 0.27$\pm$0.02          \\
                           & HITONPC       & 0.37$\pm$0.01          & 0.39$\pm$0.01 & 0.43$\pm$0.01 & 0.38$\pm$0.43          & 0.40$\pm$0.01          & 0.42$\pm$0.01 & 0.46$\pm$0.01 & 0.30$\pm$0.01          \\
                           & PC-simple     & -                  & -         & -         & -                  & -                  & -         & -         & -                  \\
                           & SLL-PC        & -                  & -         & -         & -                  & -                  & -         & -         & -                  \\
                           & S$^{2}$TMB-PC & -                  & -         & -         & -                  & -                  & -         & -         & -                  \\
                           & FCBF          & \textbf{0.47$\pm$0.01} & 0.59$\pm$0.01 & 0.43$\pm$0.01 & \textbf{0.03$\pm$0.00} & \textbf{0.49$\pm$0.00} & 0.60$\pm$0.01 & 0.45$\pm$0.00 & \textbf{0.03$\pm$0.00} \\\hline
\multirow{6}{*}{Gene}      & MMPC          & 0.83$\pm$0.00          & 0.79$\pm$0.00 & 0.92$\pm$0.00 & 0.13$\pm$0.00          & 0.84$\pm$0.00          & 0.79$\pm$0.00 & 0.94$\pm$0.00 & 0.15$\pm$0.00          \\
                           & HITONPC       & 0.83$\pm$0.00          & 0.79$\pm$0.00 & 0.92$\pm$0.00 & 0.16$\pm$0.00          & 0.84$\pm$0.00          & 0.79$\pm$0.00 & 0.94$\pm$0.00 & 0.17$\pm$0.00          \\
                           & PC-simple     & \textbf{0.97$\pm$0.00} & 0.97$\pm$0.00 & 0.97$\pm$0.00 & 0.19$\pm$0.00          & 0.97$\pm$0.01          & 0.97$\pm$0.01 & 0.98$\pm$0.00 & 0.24$\pm$0.00          \\
                           & SLL-PC        & -                  & -         & -         & -                  & -                  & -         & -         & -                  \\
                           & S$^{2}$TMB-PC & -                  & -         & -         & -                  & -                  & -         & -         & -                  \\
                           & FCBF          & \textbf{0.97$\pm$0.00} & 0.98$\pm$0.00 & 0.97$\pm$0.00 & \textbf{0.03$\pm$0.00} & \textbf{0.98$\pm$0.00} & 0.99$\pm$0.00 & 0.97$\pm$0.00 & \textbf{0.04$\pm$0.00} \\
\hline
\end{tabular}
\label{tb6-5}
\end{table*}

\begin{table*}[t]
\centering
\centering \caption{Comparison of PC learning methods with different groups of data sets in terms of normalized metrics ($\downarrow$ means the smaller, the better while $\uparrow$ denotes that the bigger, the better)}
\scriptsize
\begin{tabular}{c||cccccc|cccccc}
\hline
              & Size=500  &            &        &       &       &      & Size=1,000 &            &        &       &       &      \\
Network       & Mildew    & Hailfinder & Munin  & Pigs  & Link  & Gene & Mildew     & Hailfinder & Munin  & Pigs  & Link  & Gene \\\hline
              & \multicolumn{12}{c}{\textbf{Normalized Time $\downarrow$}}                                                                    \\
FCBF          & 1.00      & 1.00       & 1.00   & 1.00  & 1.00  & 1.00 & 1.00       & 1.00       & 1.00   & 1.00  & 1.00  & 1.00 \\
MMPC          & 5.14      & 13.48      & 66.12  & 10.26 & 14.67 & 4.35 & 10.06      & 16.26      & 96.32  & 7.54  & 9.76  & 3.71 \\
HITONPC       & 5.91      & 20.20      & 999.69 & 13.74 & 12.67 & 5.30 & 10.40      & 19.02      & 550.14 & 8.56  & 10.92 & 4.27 \\
PC-simple     & 294.17    & 16.93      & -      & 18.32 & -     & 6.38 & 180.58     & 16.76      & -      & 13.54 & -     & 6.62 \\
SLL-PC        & 141780.00 & 3894.44    & -      & -     & -     & -    & -          & 2060.31    & -      & -     & -     & -    \\
S$^{2}$TMB-PC & 120580.00 & 3866.67    & -      & -     & -     & -    & -          & 1672.26    & -      & -     & -     & -    \\\hline
              & \multicolumn{12}{c}{\textbf{Normalized F1 $\uparrow$}}                                                                      \\
FCBF          & 1.00      & 1.00       & 1.00   & 1.00  & 1.00  & 1.00 & 1.00       & 1.00       & 1.00   & 1.00  & 1.00  & 1.00 \\
MMPC          & 0.90      & 0.92       & 0.36   & 0.95  & 0.79  & 0.86 & 0.93       & 1.00       & 0.49   & 0.93  & 0.82  & 0.86 \\
HITONPC       & 0.86      & 1.00       & 0.36   & 0.95  & 0.79  & 0.86 & 0.68       & 1.00       & 0.51   & 0.93  & 0.82  & 0.86 \\
PC-simple     & 0.46      & 1.00       & -      & 1.03  & -     & 1.00 & 0.43       & 1.00       & -      & 1.01  & -     & 0.99 \\
SLL-PC        & 0.90      & 0.92       & -      & -     & -     & -    & -          & 1.08       & -      & -     & -     & -    \\
S$^{2}$TMB-PC & 0.92      & 0.92       & -      & -     & -     & -    & -          & 1.08       & -      & -     & -     & -    \\\hline

\end{tabular}
\label{tb6-6}
\end{table*}

\subsubsection{Experimental analysis of the threshold $\delta$ for FCBF}

Although FCBF does not require users to specify the number of selected features before learning, it needs a user-specified parameter $\delta$ to control the number of candidate parents and children of a variable of interest at the forward step of FCBF as described in Section~\ref{sec43}.  A large $\delta$ makes FCBF remove  false positives as well as  true positives. A small $\delta$ leads FCBF to find more true positives and more false positives as well.

To choose an optimal value of $\delta$  for  FCBF, we have investigated the F1, recall, and precision of FCBF using different $\delta$ value from 0 to 0.1 with the six BNs in Table~\ref{tab6-0}. In Figure~\ref{sp-1}, we can see that when $\delta$ is small, FCBF gets a low value of F1. As $\delta$ increases, the value of F1 gradually becomes high and stable.
The explanation is that when $\delta$ is small, as shown in Figure~\ref{sp-3}, many false positives will enter the final output of FCBF, this reduces the precision of FCBF. However, Figure~\ref{sp-4} illustrates that the value of $\delta$ does not have much impact on the true PC entering the output of FCBF, since parents and children of a variable should have a strong dependency relationship  with the variable and they are less likely be discarded even when $\delta$ becomes big. As  $\delta$ increases, some true positives will be discarded.
Using the six BNs, we can see that when $\delta$ is up to 0.05, FCBF is stable and almost gets the highest  F1, and thus in all the experiments in Section~\ref{sec6}, we choose 0.05 as the value of  $\delta$ for FCBF.

\begin{figure*}[t]
\centering
       \begin{tabular}{ccccc}
   \subfigure{\includegraphics[width=2.8in, height=1.7in]{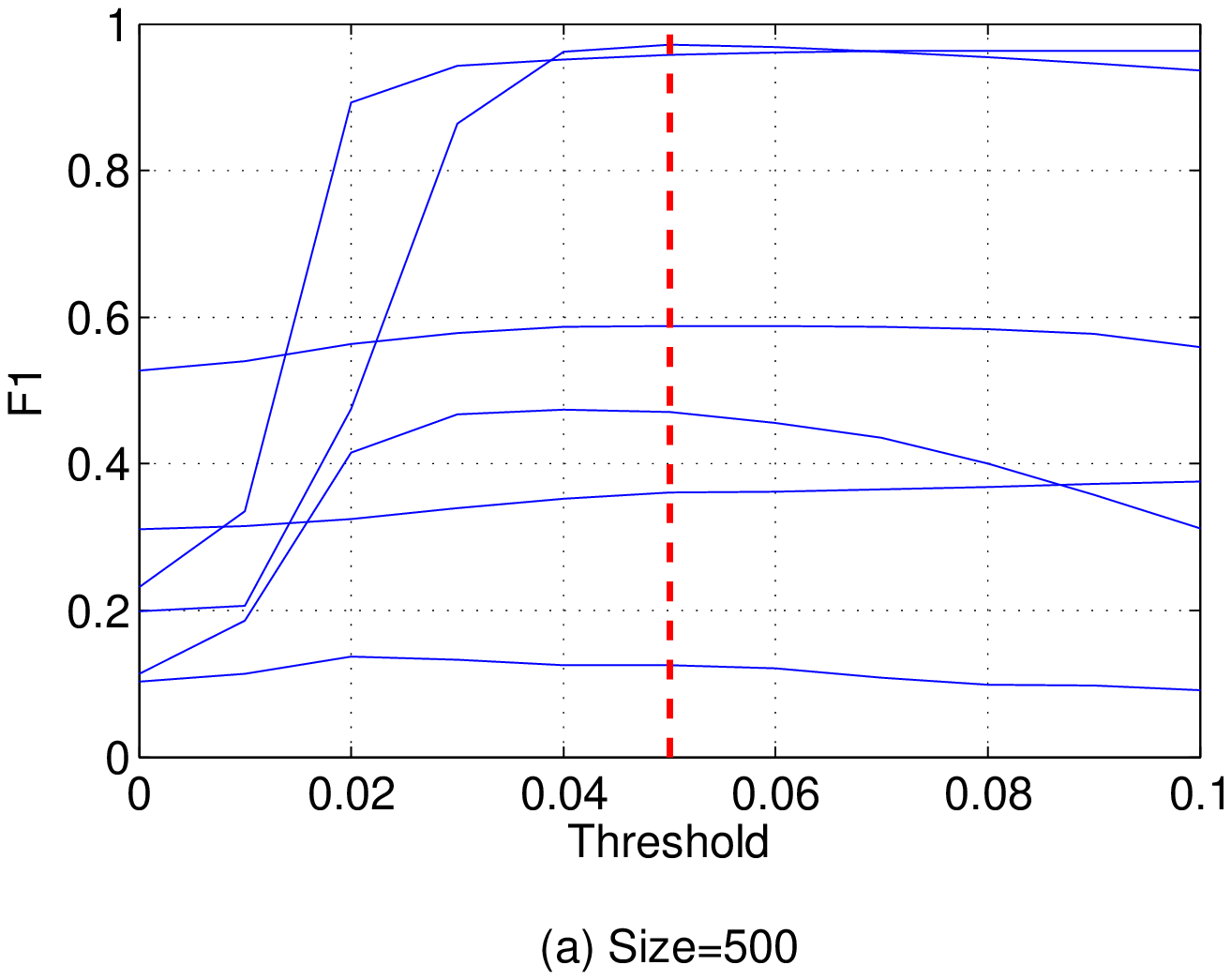}}&
  \subfigure{\includegraphics[width=2.8in, height=1.7in]{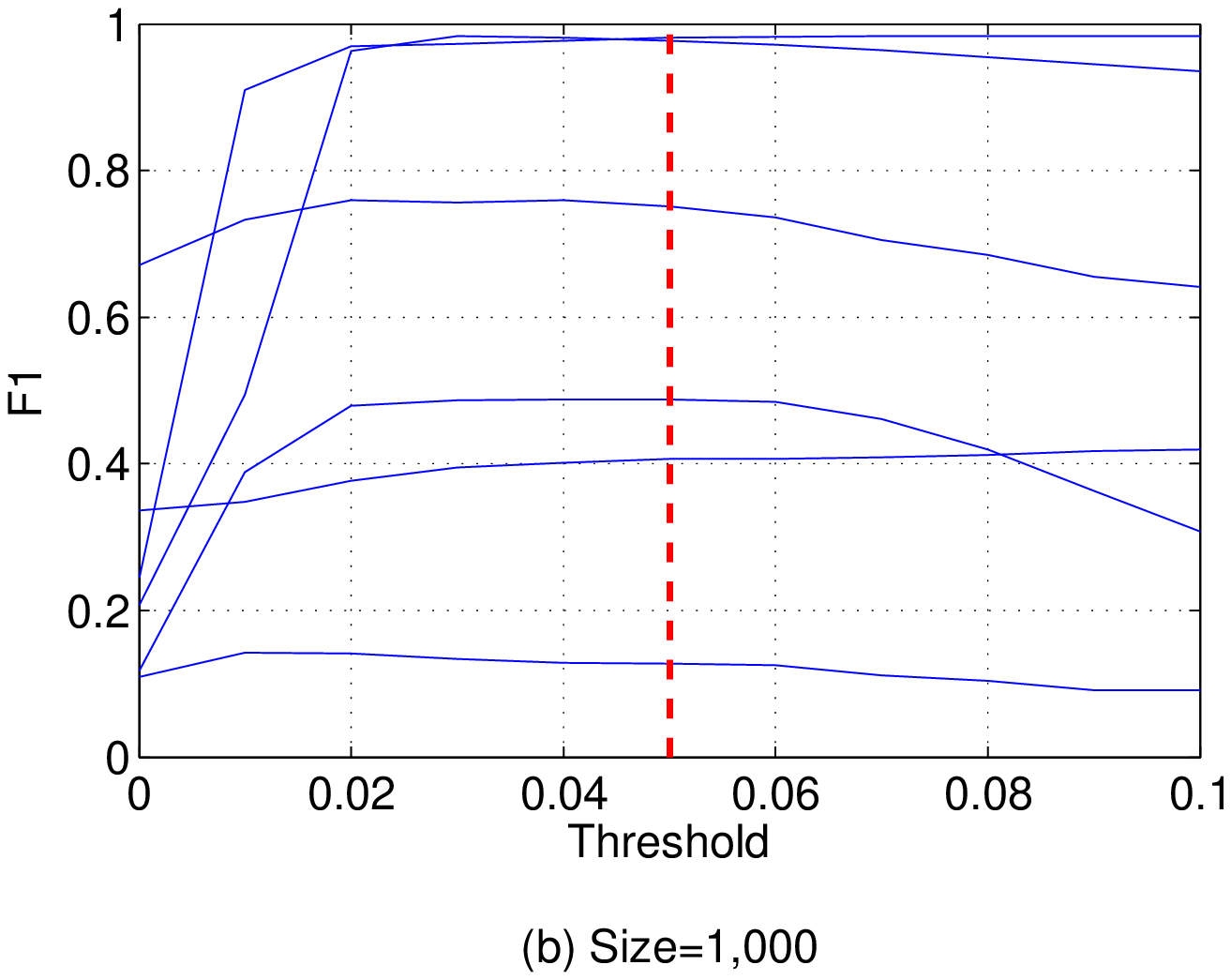}}&
      \end{tabular}
 \caption{F1 of FCBF on the six BNs when varying the value of   $\delta$ from 0 to 0.1 with different groups of data sets. The red  line denotes $\delta$=0.05.}
 \label{sp-1}
\end{figure*}

\begin{figure*}[t]
\centering
       \begin{tabular}{ccccc}
      \subfigure{\includegraphics[width=2.8in, height=1.7in]{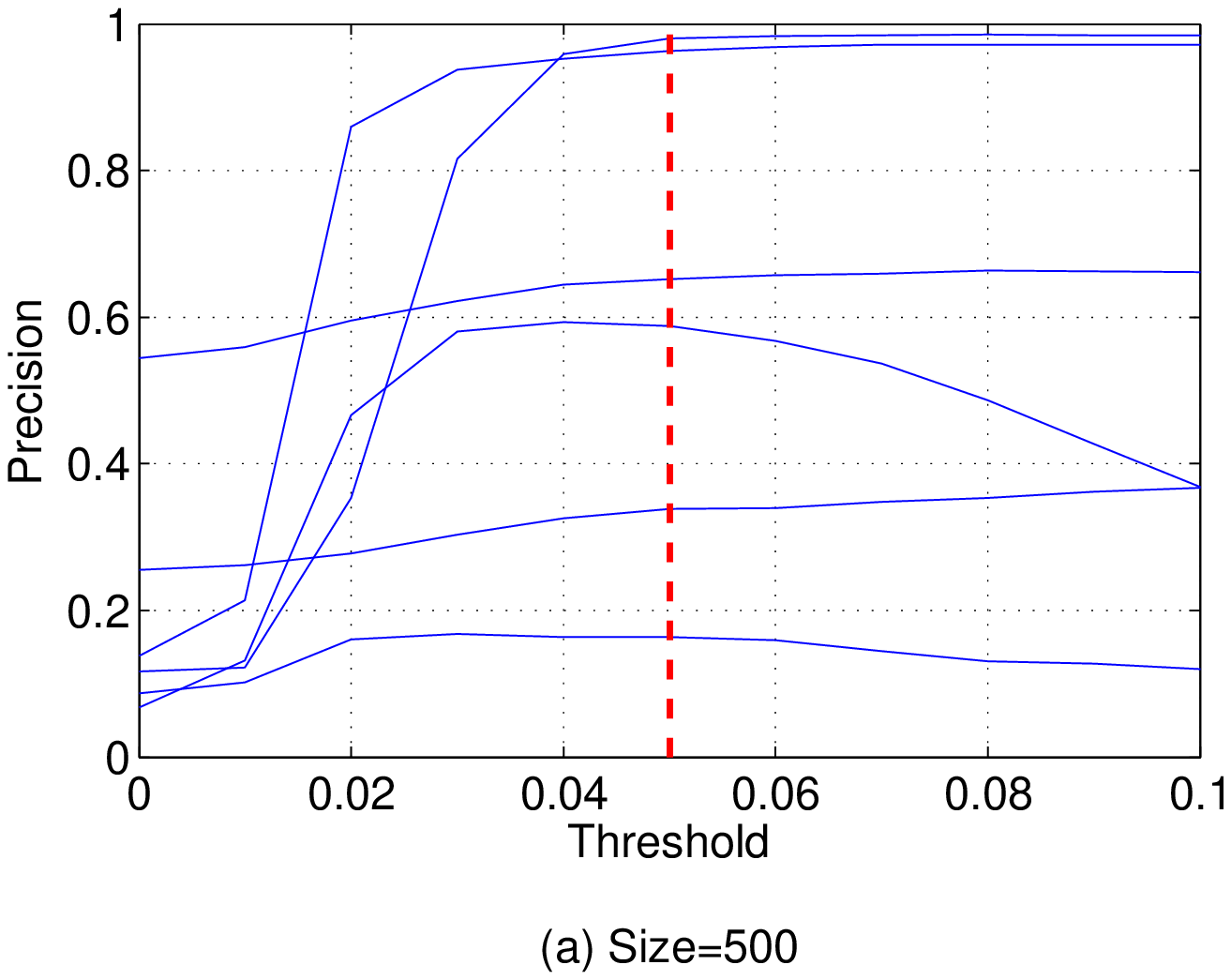}}&
      \subfigure{\includegraphics[width=2.8in, height=1.7in]{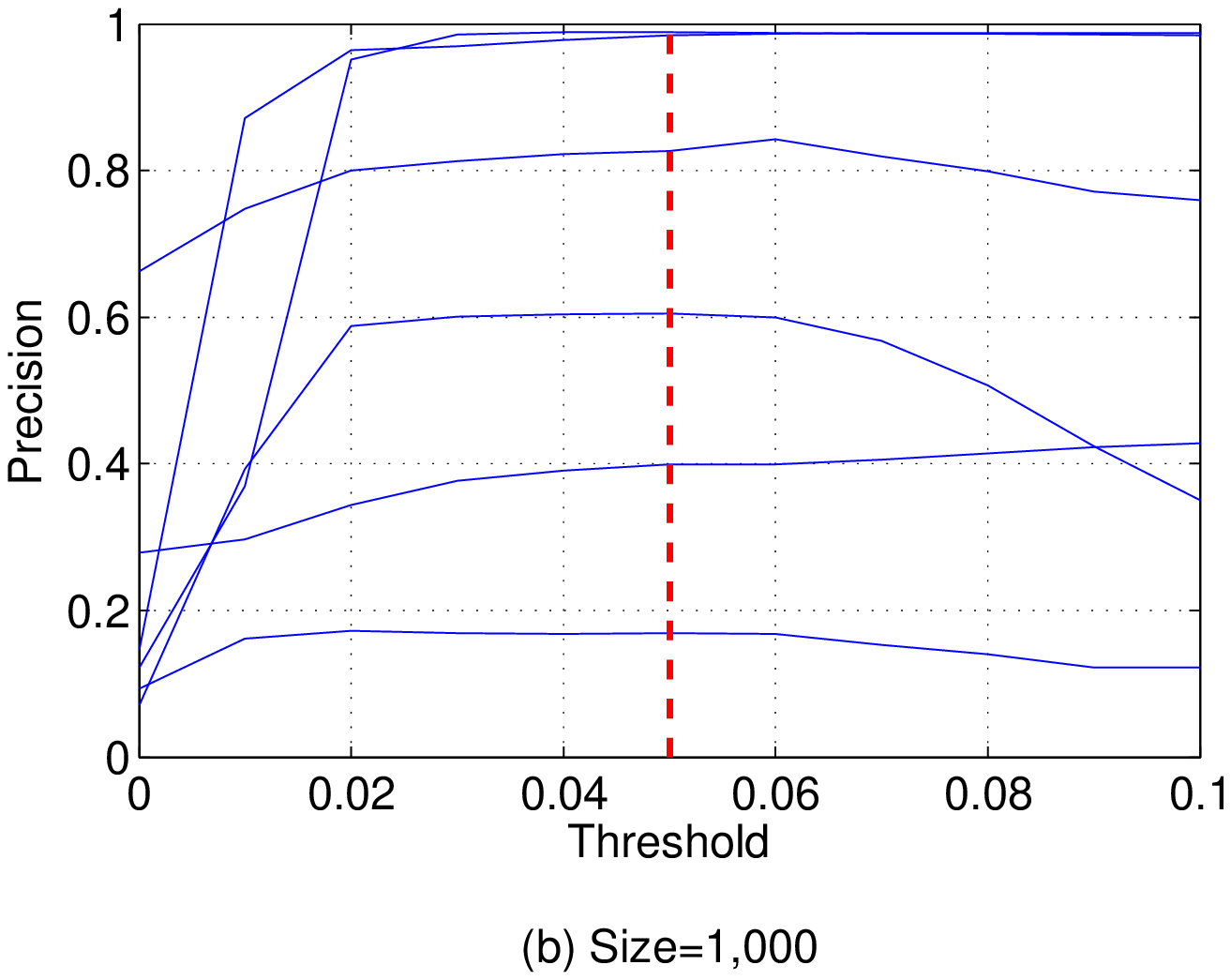}}&
       \end{tabular}
 \caption{Precision of FCBF on the six BNs when varying the value of   $\delta$ from 0 to 0.1 with different groups of data sets. The red  line denotes $\delta$=0.05.}
 \label{sp-3}
\end{figure*}

\begin{figure*}[t]
\centering
       \begin{tabular}{ccccc}
      \subfigure{\includegraphics[width=2.8in, height=1.8in]{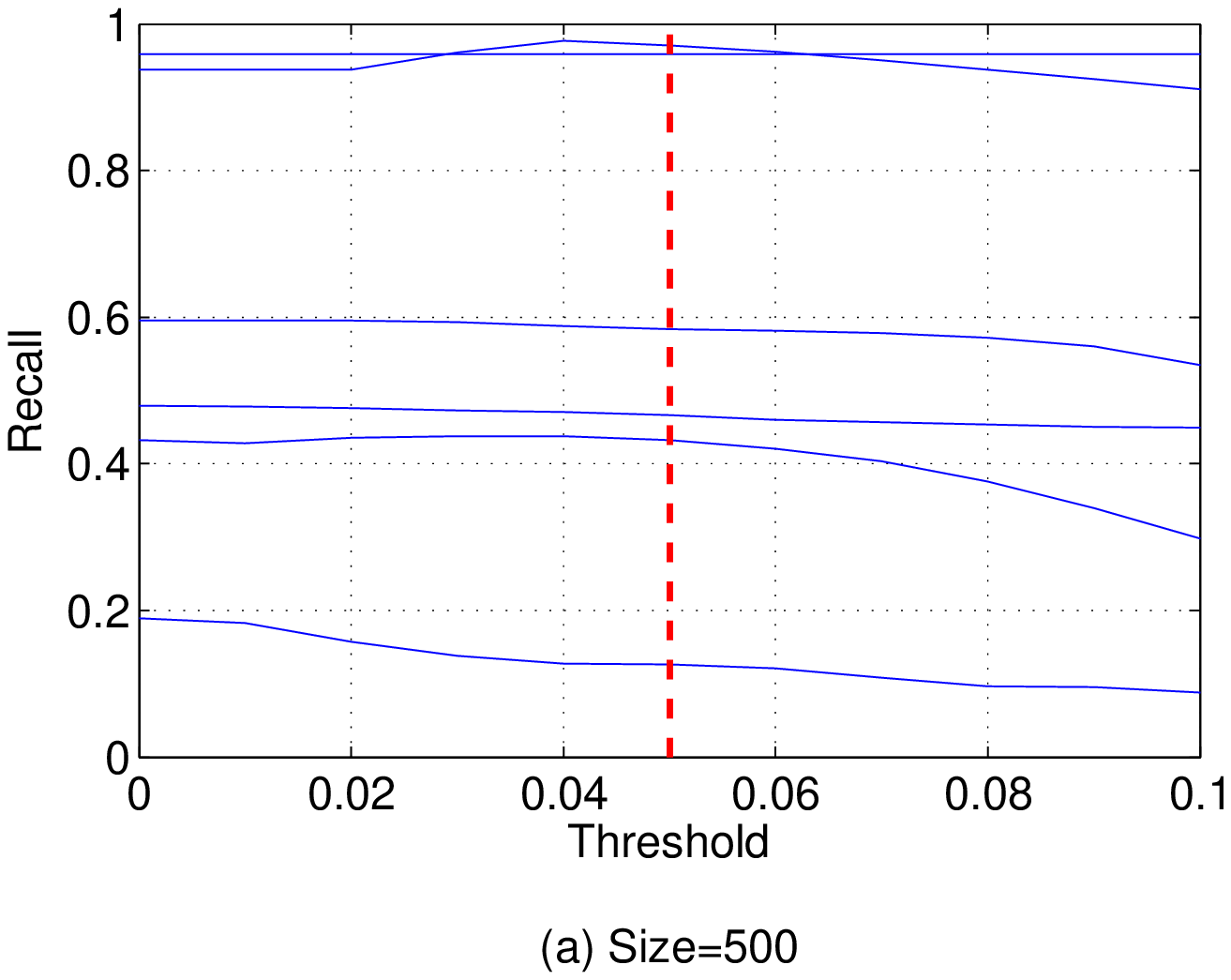}}&
      \subfigure{\includegraphics[width=2.8in, height=1.8in]{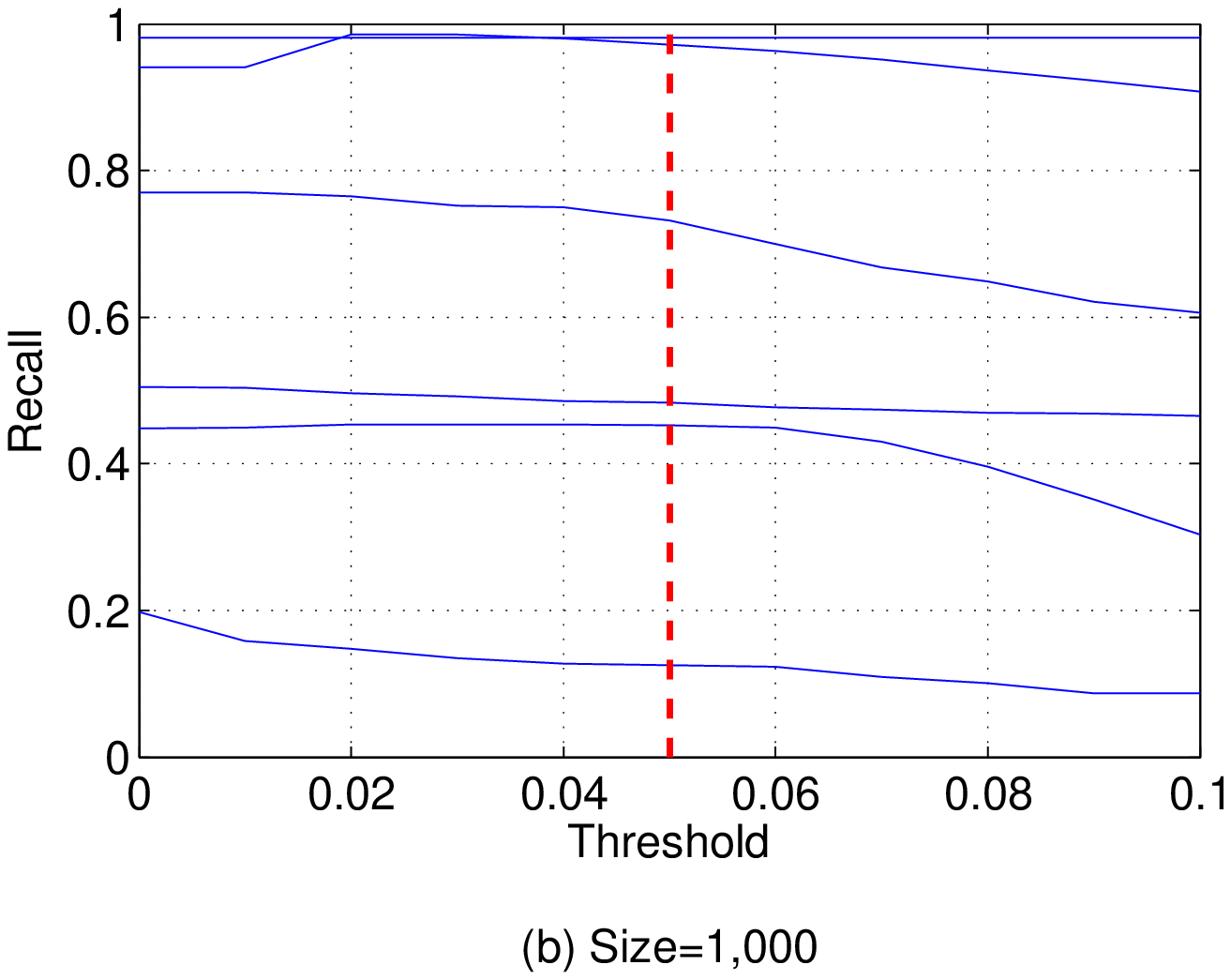}}&
       \end{tabular}
 \caption{Recall of FCBF on the six BNs when varying the value of   $\delta$ from 0 to 0.1 with different groups of data sets. The red  line denotes $\delta$=0.05.}
 \label{sp-4}
\end{figure*}

\section{Conclusion}\label{sec7}

In this paper, we link feature selection methods to BN structure learning to improve computational efficiency, and propose the F2SL framework for BN structure learning by using the FCBF algorithm. By instantiating the F2SL framework, we propose two efficient local-to-global structure learning algorithms, F2SL-c and F2SL-s. Using six benchmark BNs, the experimental results have shown that F2SL-c and F2SL-s significantly improve computational efficiency of BN structure learning compared to the eight state-of-the-art BN structure learning algorithms and  both also achieve competitive structure learning quality with the eight rivals.

\section*{Acknowledgments}
{
This work is partially supported by the National Key Research and Development Program of China (under grant 2020AAA0106100),  National Natural Science Foundation of China (under Grant 61876206), and Open Project Foundation of Intelligent Information Processing Key Laboratory of Shanxi Province (under grant CICIP2020003).
}

\bibliographystyle{abbrv}

\bibliography{new}

\end{sloppy}
\end{document}